\documentclass{article}

\usepackage{arxiv}

\usepackage{makecell, tabularx}
\usepackage[export]{adjustbox}
\usepackage[most]{tcolorbox} 
\usepackage{algorithm}
\usepackage{tabularx}
\usepackage{rotating}
\usepackage[algo2e]{algorithm2e} 
\usepackage{algpseudocode}
\usepackage{url}
\usepackage[table,xcdraw]{xcolor}
\usepackage{hyperref}
\usepackage[numbers]{natbib}
\usepackage{amsmath,amssymb}
\usepackage{bbm}
\usepackage{soul}
\usepackage[utf8]{inputenc}
\usepackage[english]{babel}
\usepackage{graphicx}
\usepackage{booktabs}
\usepackage{multirow}
\usepackage{subcaption}
\usepackage{algpseudocode}
\usepackage{algorithm}
\usepackage{minted}
\usepackage{tabularray}
\usepackage{pifont}

\usepackage[font=footnotesize]{caption}  
\newcolumntype{C}[1]{>{\centering\arraybackslash}p{#1}}
\renewcommand{\shorttitle}{SustainFM}

\usepackage{authblk}

\author[1,2]{Pedram Ghamisi} 
\author[1]{Weikang Yu}
\author[3]{Xiaokang Zhang}
\author[1]{Aldino Rizaldy}
\author[4]{Jian Wang}
\author[4]{Chufeng Zhou}
\author[1]{Richard~Gloaguen}
\author[5]{Gustau Camps-Valls}

\affil[1]{Helmholtz-Zentrum Dresden-Rossendorf (HZDR), 09599 Freiberg, Germany} 
\affil[2]{University of Iceland, 101 Reykjavík, Iceland} 
\affil[3]{Wuhan University, 430072 Wuhan, China} 
\affil[4]{Wuhan University of Science and Technology, 430081 Wuhan, China}
\affil[5]{Universitat de València, 46010 València, Spain}

\date{}    
\title{Geospatial Foundation Models to Enable Progress on Sustainable Development Goals}

\begin{document}
\maketitle
\begin{abstract}
Foundation Models (FMs) are large-scale, pre-trained artificial intelligence (AI) systems that have revolutionized natural language processing and computer vision, and are now advancing geospatial analysis and Earth Observation (EO). They promise improved generalization across tasks, scalability, and efficient adaptation with minimal labeled data. However, despite the rapid proliferation of geospatial FMs, their real-world utility and alignment with global sustainability goals remain underexplored. 
We introduce SustainFM, a comprehensive benchmarking framework grounded in the 17 Sustainable Development Goals with extremely diverse tasks ranging from asset wealth prediction to environmental hazard detection. This study provides a rigorous, interdisciplinary assessment of geospatial FMs and offers critical insights into their role in attaining sustainability goals.
Our findings show: (1) While not universally superior, FMs often outperform traditional approaches across diverse tasks and datasets. (2) Evaluating FMs should go beyond accuracy to include transferability, generalization, and energy efficiency as key criteria for their responsible use. (3) FMs enable scalable, SDG-grounded solutions, offering broad utility for tackling complex sustainability challenges.
Critically, we advocate for a paradigm shift from model-centric development to impact-driven deployment, and emphasize metrics such as energy efficiency, robustness to domain shifts, and ethical considerations.
\end{abstract}


\begin{figure*}[t]
	\centering
	\includegraphics[width=1\textwidth]{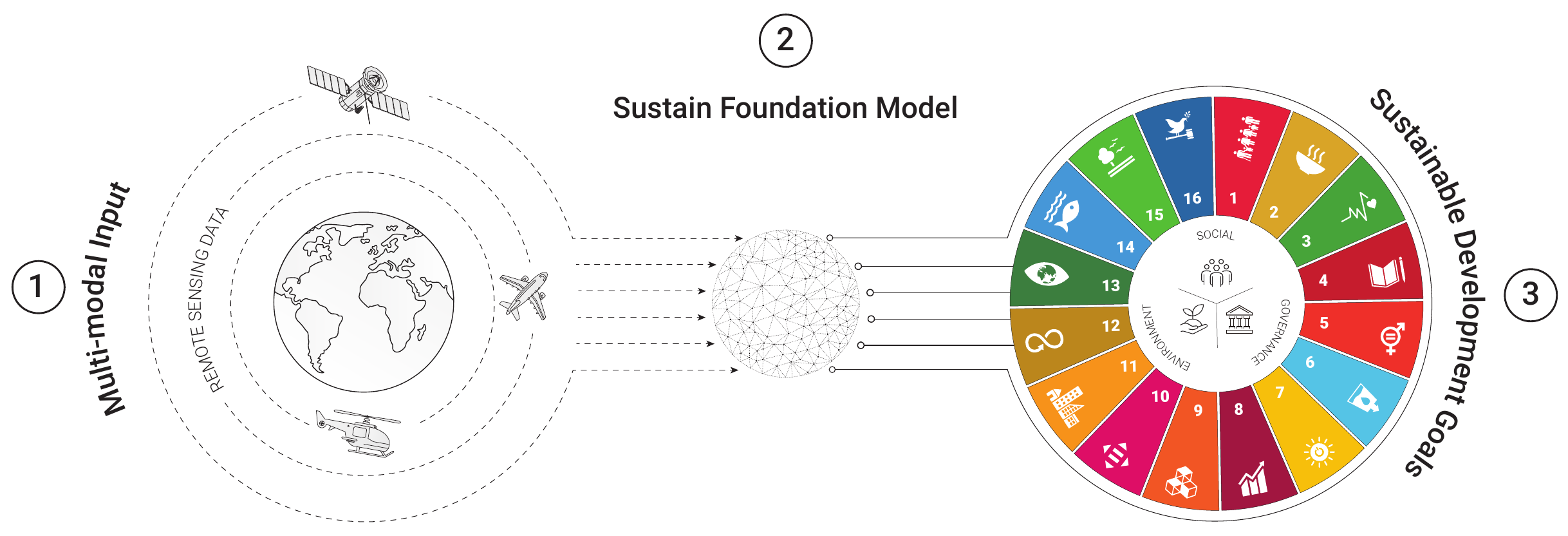}
    \vspace{-8pt}
	\caption{Summary of the SustainFM benchmark. By bringing together domain experts and AI specialists (SDG 17: Partnerships for the Goals), and providing a collection of 16 datasets aligned with the first 16 SDGs, from six continents, spatial resolutions ranging from 0.5 m to 30 m, and multiple tasks grounded in real-world applications, SustainFM aims to provide a testbed to analyze the applicability and real impact of foundation models.}
	\label{fig:main}
\end{figure*}

\section{Introduction}
The convergence of artificial intelligence (AI) and Earth observation (EO) technologies has launched an unprecedented era of capabilities for addressing global sustainability challenges \cite{burke2021using}. The Sustainable Development Goals (SDGs), established by the United Nations General Assembly in 2015, represent a comprehensive framework for addressing global challenges through 17 interconnected goals spanning social, economic, and environmental dimensions \cite{united2017sustainable,sachs2019six,hak2016sustainable}. These goals aim to achieve a better and more sustainable future by 2030, addressing critical issues including poverty, inequality, climate change, environmental degradation, peace, and justice \cite{sorooshian2024sustainable,kavvada2020towards,9681713}. EO techniques significantly contribute to sustainable development by providing critical data and insights that support the achievement of SDGs \cite{kavvada2020towards,9681713, vinuesa2020role}. From monitoring climate change, biodiversity, and water resources to enhancing food security, urban planning, and disaster resilience \cite{nature_camps2025artificial, nature_tamayo2025advanced,nature_ratledge2022using,nature_hughes2024integrating}, EO provides critical information for tracking environmental and socio-economic trends. Through supporting evidence-based policy and providing an enormous amount of up-to-date data, EO accelerates progress toward a more sustainable, equitable, and resilient future \cite{zhao2025advancing,ferreira2020monitoring}. 

% EO technologies provide critical support for SDG monitoring and implementation through their systematic data collection and analysis capabilities, enabling evidence-based decision-making \cite{zhao2025advancing,ferreira2020monitoring}. 

The scale and complexity of EO data necessitate automated analytical tools capable of handling diverse, multimodal spatio-temporal datasets \cite{ghamisi2019multisource}.  Foundation models (FMs) have emerged as a transformative solution, offering unprecedented capabilities for geospatial applications through their ability to process vast amounts of heterogeneous data and generalize across multiple downstream tasks \cite{zhang2024geoscience,xiao2025foundation}.
FMs represent a paradigm shift in AI, characterized by large-scale pre-training on extensive, diverse datasets using self-supervised \cite{xu2025sdcluster} or fully supervised learning approaches \cite{wang2022advancing}. These models normally contain billions of parameters (e.g., Google’s SegGPT \cite{wang2023seggpt} with over 3 billion and Meta’s DINOv2 \cite{oquab2023dinov2} with around 1 billion parameters), and leverage vast computational resources to learn detailed patterns and representations from massive datasets, which enables them to generalize across a wide range of downstream tasks. 
With this definition, they serve as a universal backbone to support a wide range of applications across domains such as natural language processing (NLP), computer vision, and (geo)scientific computing. The term foundation emphasizes its role in providing a pre-trained, transferable representation that can be fine-tuned or adapted for domain-specific applications. Specifically, geospatial FMs are pre-trained on vast amounts of airborne and spaceborne data and can then be adapted for downstream EO tasks, such as land cover classification, change detection, or disaster response mapping, without requiring separate models to be trained from scratch for each task  \cite{lu2025vision}.
Key characteristics of FMs include: (1) \textbf{Scalability} through efficient, distributed training; (2) \textbf{Generalization} via transferable representations across tasks; and (3) \textbf{Few-/Zero-shot Learning}, enabling adaptation with little or no labeled data.

The evolution of geospatial FMs has witnessed explosive growth, progressing from single-modal approaches focusing on optical satellite imagery \cite{hong2024spectralgpt,sun2022ringmo,li2024s2mae} to sophisticated multi-modal architectures that integrate diverse data sources \cite{guo2024skysense,zhang2024earthgpt,nedungadi2024mmearth, li2025fleximo}, including optical, SAR, hyperspectral, and auxiliary geospatial information \cite{10766851,wu2025semantic}. 
However, the rapid growth in the number of FMs over such a short period has led to a saturation of new architectures, resembling the clear trajectory we witnessed in the EO community during the deep learning boom around 2016, often marked by incremental improvements and limited consideration of their societal, environmental,
and economical impact.
A fundamental gap remains in grounding model development with global
sustainability objectives. To see the actual impact of FMs on real-world challenges, such as climate monitoring, resource management, and humanitarian applications, there needs to be a stronger focus on the deployment phase (downstream tasks), which is widely regarded as the ultimate goal of Responsible AI in EO \cite{ghamisi2025responsible}.

On the other hand, current evaluation approaches primarily focus on task-specific accuracy metrics \cite{lacoste2023geo,dionelis2024evaluating,marsocci2024pangaea}, providing limited insight into the model's utility for addressing complex sustainability challenges. This gap is particularly concerning when FMs are increasingly applied in domains directly related to human welfare, environmental conservation, and economic development areas, where decisions carry profound consequences beyond accuracy scores. To truly unlock their potential, FMs should be assessed not only on technical sophistication but also on their contributions to social equity, environmental protection, and economic sustainability. Yet, such dimensions are often absent in current evaluation frameworks, underscoring the urgent need for a systematic benchmark that integrates mainstream FMs with multi-source datasets aligned to the SDGs.

% Current evaluation practices for geospatial Foundation Models (FMs) remain largely confined to computational metrics and academic benchmarks \cite{dionelis2024evaluating,eriksson2025can}, leading to a clear disconnect between technical performance claims and their actual societal impact. 

% Current evaluation approaches focus primarily on task-specific accuracy metrics, providing limited insight into model utility for addressing complex sustainability challenges. 
% For SDGs, comprehensive evaluation frameworks must incorporate multiple dimensions beyond accuracy, including transferability across geographic regions and tasks, data efficiency in few-shot learning scenarios, computational costs and environmental impact (energy consumption, CO2 emissions), operational effectiveness in addressing practical challenges, and alignment with sustainability objectives.

To address this, we must shift the emphasis from mere model development to the deployment phase \cite{radanliev2024ethics}, where FMs are evaluated based on their impact on complex socio-environmental contexts. Performance evaluation should go beyond traditional benchmarks and include real-world scenarios, social good, and ethical considerations in the deployment stage. Furthermore, the interdisciplinary views of both AI and domain experts are required to ensure that FMs are not only optimized for accuracy but also for their contribution to long-term societal benefits  \cite{tassa2022going}. Without this shift, we risk an unsustainable model development cycle with no real-world impacts.

In this paper, we move beyond the current hype of FMs to adopt an evidence-based approach to answer this question:

\begin{tcolorbox}[
    enhanced,
    colback=gray!8,      % light background
    colframe=gray!60,    % border colour
    boxrule=0.8pt,       % border thickness
    arc=4pt,             % rounded corners
    left=6pt,            % inner left padding
    right=6pt,           % inner right padding
    top=6pt,             % inner top padding
    bottom=6pt           % inner bottom padding
]
\textit{1. Do we need FMs, or are traditional, purpose-specific deep learning models sufficient for EO tasks?}\\[0.6em]
\textit{2. What evaluation criteria best capture the effectiveness and responsibility of FMs in EO, beyond task-specific accuracy?}\\[0.6em]
\textit{3. Can FMs be systematically leveraged to support scalable, SDG-grounded solutions for geospatial challenges?}
\end{tcolorbox}

Through systematic evaluation across key criteria, including accuracy, transferability, data efficiency, energy costs, and operational impact, we aim to provide a grounded, rigorous assessment that challenges assumptions and clarifies the actual added value, if any, of FMs in this domain. Ultimately, our goal is to contribute to global sustainability objectives and promote social good, a key aspect that has been undermined in many current geospatial FMs. To evaluate FMs in EO rigorously, we introduce SustainFM, a benchmark aligned with all 17 SDGs that provides new EO datasets, leverages pretrained FMs, and evaluates them for global sustainability, as conveyed in Fig. \ref{fig:main}.

% This dataset spans a wide range of geographies (over 200 countries and regions across continents, as shown in global distribution in Fig. \ref{fig:map}), sensor types, and spatial resolutions (from 0.5 m to 30 m) to reflect the diversity and complexity of global sustainability challenges. SustainFM combines 16 real-world tasks (dedicated to the first 16 SDGs), including asset wealth prediction, child and women's health estimation, solar farm and settlement mapping, and detecting industrial smoke, marine pollution, and conflict-related damage. These tasks cover key learning tasks such as regression, change detection, segmentation, and classification, and are tackled using a variety of sensors such as Landsat, Sentinel-1/2, VIIRS, Gaofen-2, Harmonized Landsat and Sentinel-2 (HLS), Google Earth, and PlanetScope. 
% By encompassing domains like climate resilience, food security, infrastructure monitoring, and humanitarian response, SustainFM enables a comprehensive assessment of FMs across transferability, data efficiency, and real-world utility. By bringing together experts with diverse backgrounds, from domain scientists and remote sensing specialists to AI ethicists, machine learning researchers, and statisticians, we contribute to SDG-17 (Partnerships for the Goals) and offer critical yet comprehensive insights into whether FMs can contribute to global sustainability, moving beyond academic benchmarks to real-world impact.

This paper makes four key contributions. Scientifically, it introduces SustainFM, the first comprehensive benchmarking framework for evaluating FMs against the SDGs, setting new standards for responsible AI in Earth observation. Methodologically, it advances beyond accuracy-focused assessments by integrating critical metrics such as energy efficiency, transferability, and environmental impact, promoting a paradigm shift from performance-driven development to impact-driven deployment. Societally, it demonstrates how FMs can enable scalable solutions to global challenges while ensuring equitable outcomes across diverse communities. Environmentally, it quantifies the carbon footprint of model training and advocates for energy-efficient strategies that reduce emissions without sacrificing performance. By aligning geospatial AI with the SDG framework, this work lays a crucial foundation for leveraging advanced AI responsibly to address pressing sustainability challenges.

\begin{figure*}[t]
	\centering
	\includegraphics[width=1\textwidth]{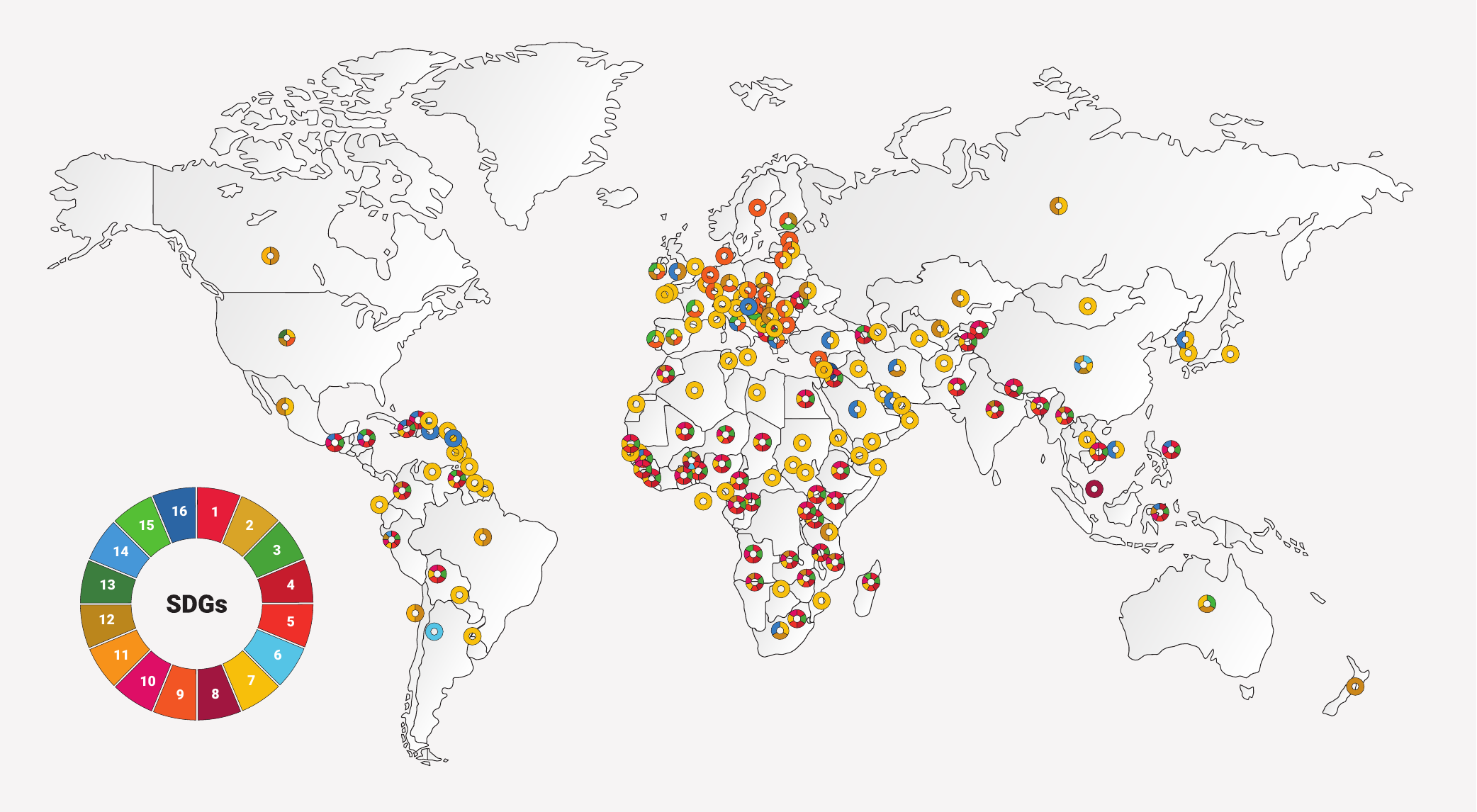}
     \vspace{-8pt}
	\caption{Geographic coverage of the SustainFM benchmark, which spans over 200 regions across 6 continents to support a wide range of EO tasks aligned with the SDGs.}
	\label{fig:map}
\end{figure*}

\begin{table*}
%\captionsetup{font=small}
\caption{Description of the SustainFM Benchmark, including SDG-related tasks (applications) with sensor type, resolution, scale, AI task, patch size, and number of patches.}
\centering
\label{tab:sustainfmbenchmark}
\resizebox{\linewidth}{!}{
\begin{tabular}{c|ccccccc}
\toprule
\textbf{\#SDG} & \textbf{Application}                                  & \textbf{Sensor}                    & \textbf{Resolution} & \textbf{Scale}    & \textbf{AI Task} & \textbf{Patch Size}       &\textbf{\#Patches}                                \\ \midrule
\rowcolor{gray!10}  \textbf{1}     & Asset Wealth Index Regression                & Landsat+VIIRS             & 30m        & Africa   & Reg.    & $256 \times 256$ &  86,936      \\
\textbf{2}     & Cropland Change Detection                    & Gaofen-2                  & 0.5-2m     & China    & CD      & $256 \times 256$ & 2,400             \\
\rowcolor{gray!10}  \textbf{3}     & Children's Health Regression              & Landsat+VIIRS             & 30m        & Africa   & Reg.    & $256 \times 256$ & 105,582    \\
\textbf{4}     & Women Educational Attainment Regression      & Landsat+VIIRS             & 30m        & Africa   & Reg.    & $256 \times 256$ &    117,062    \\
\rowcolor{gray!10}  \textbf{5}     & Women Health Regression                         & Landsat+VIIRS             & 30m        & Africa   & Reg.    & $256 \times 256$ &   94,866      \\
\textbf{6}     & Urban BOW Detection                          & Gaofen-2                  & 1m         & China    & Seg.    & $256 \times 256$ &      1,645        \\
\rowcolor{gray!10}  7     & Solar Farm Identification                    & Sent-2                & 10m        & Global   & Seg.    & $256 \times 256$ &   10,230       \\
\textbf{8}     & Detection of Settlements without Electricity & Sent-1+Sent-2+Landsat+VIIRS & 10m        & Malawi   & Seg.    & $800 \times 800$ &   98           \\
\rowcolor{gray!10}  \textbf{9}     & Industrial Smoke Plume Detection             & Sent-2                & 10m        & EU       & Seg.    & $120 \times 120$ & 25,100 \\
\textbf{10}    & Sanitation Index Regression                  & Landsat+VIIRS             & 30m        & Africa   & Reg.    & $256 \times 256$ &     89,271   \\ 
\rowcolor{gray!10} \textbf{11}    & Urban Building Change Detection              & Google Earth              & 0.5m       & U.S.     & CD      & $1024 \times 1024$ &   637        \\
\textbf{12}    & Mining Change Detection                      & Google Earth              & 1.2m       & Global   & CD      & $256 \times 256$ &      71,711            \\
\rowcolor{gray!10} \textbf{13}    & Wildfiire Burn Scar Detection                & HLS                       & 30m        & U.S.     & Seg.    & $512 \times 512$ &  1,068    \\
\textbf{14}    & Marine Pollution Detection                   & Sent-2                & 10m        & Global   & Seg.    & $240 \times 240$ &   2,803         \\
\rowcolor{gray!10} \textbf{15}    & Flood Mapping                                & Sent-1+Sent-2     & 10m        & Global   & Seg.    & $256 \times 256$ &      1,688                \\
\textbf{16}    & War-induced Damage Classification            & Planet Scope              & 3m         & Regional & Cls.    & $128 \times 128$ &   13,460   \\
%\rowcolor{gray!10}  \textbf{17}    &                                              &                           &            &          &         &         &                                     \\ 
\bottomrule
\end{tabular}}
\end{table*}

\section{SustainFM}
Achieving the SDGs requires reliable tools to monitor progress across complex, interconnected systems. Despite the global availability of satellite data, assessing the real-world impact of development processes and their alignment with sustainability goals remains a major challenge \cite{9681713}. To address this gap, we introduce SustainFM, a benchmark dataset for monitoring $16$ SDGs using EO data.
Specifically, SustainFM spans over 200 countries and regions across continents (Fig. \ref{fig:map}), covering diverse sensor types and spatial resolutions ($0.5$ m–$30$ m), including Landsat, Sentinel-1/2, VIIRS, Gaofen-2, HLS, Google Earth, and PlanetScope, to reflect the complexity of global sustainability challenges.
It integrates 16 real-world tasks aligned with the first 16 SDGs, including asset wealth prediction, health estimation, solar farm and settlement mapping, as well as detection of industrial smoke, marine pollution, and conflict-related damage. These tasks represent key learning paradigms, including regression (Reg.), classification (Cls.), semantic segmentation (Seg.), and change detection (CD). 
% By encompassing domains such as climate resilience, food security, infrastructure monitoring, and humanitarian response, SustainFM enables comprehensive evaluation of FMs in terms of transferability, data efficiency, and real-world utility. 
Through collaboration with experts from diverse fields, including remote sensing, machine learning, statistics, and AI ethics, SustainFM also contributes to SDG-17 (Partnerships for the Goals), bridging the gap between academic benchmarks and real-world impact. Table \ref{tab:sustainfmbenchmark} summarizes the EO datasets in SustainFM, with each entry corresponding to a task linked to a specific SDG application. 
Designed as a testbed for downstream tasks, SustainFM supports systematic evaluation of FM performance, generalization, and energy efficiency across data modalities and learning settings. 
% It minimizes regional bias through its global coverage and addresses real-world challenges spanning health, education, infrastructure, industrial activity, and disaster response. 
Below, we introduce the 16 building blocks of SustainFM, each aligned with a specific SDG, along with the corresponding evaluation protocols.

\subsection{No Poverty (SDG-1)} 
%Sub-dataset 1 SDG1
Eradicating poverty in all its forms is a central objective of sustainable development. 
% Poverty is closely linked to a lack of access to resources, infrastructure, and basic services such as education and healthcare.
Effective monitoring of poverty requires capturing spatial and socioeconomic disparities, which can be challenging using traditional survey methods alone.  EO data offers a complementary approach by providing large-scale insights into analyzing poverty triggers from a bird's-eye view of residential conditions, including human settlements, infrastructure quality, and resource distribution \cite{van2021satellite,cigna2022urban,pesaresi2024advances}. 
% Although the EO data has been widely investigated to map the land use and land cover specifications that benefit a wide range of SDGs, 
However, human development circumstances within society, such as human well-being, education, and gender equality, still face difficulties in being directly measured or indicated by EO data due to the lack of labels. 

To tackle these challenges, recent literature has seen innovations that indicate the development of such themes by using human survey data and subsequently connecting these survey data with satellite images via regression tasks \cite{brown2014using,sharma2024kidsat}. One possible solution is to use the Demographic and Health Surveys (DHS) survey data, which has been widely explored for its relationship with the satellite images. For example, SustainBench \cite{yeh2021sustainbench} explored six variables in the DHS dataset, including asset wealth index, women's BMI, child mortality rate, women's educational attainment, water quality index, and sanitation index. In particular, the SustainBench processes the DHS survey data by converging the localized entries into cluster-level data, and subsequently collects satellite images that are geographically centered around each surveyed cluster’s geocoordinates. These satellite images comprise both daytime images from the Landsat 5, 7, and 8 satellites and nightlight images from the DMSP and VIIRS satellites.

Therefore, SustainFM includes an asset wealth index regression task to assess localised poverty using the DHS data from SustainBench \cite{yeh2021sustainbench}. Wealth is estimated at the cluster level using a principal component analysis (PCA) of household assets, yielding a scalar index averaged across surveyed households in each cluster. The dataset comprises $2,079,036$ households across $86,936$ clusters in $48$ countries, based on DHS surveys conducted between 1996 and 2019. 

\subsection{Zero Hunger (SDG-2)} %SDG2
Ensuring food security and promoting sustainable agriculture is critical to combating hunger and malnutrition. Agricultural productivity is influenced by factors such as weather conditions, soil health, and water availability, all of which can be monitored using EO data. Satellite imagery enables large-scale assessment of cropland distribution, yield estimation, and the impact of natural disaster incidents \cite{joyce2009review,jiang2020large}. Within these perspectives, monitoring the fast and dynamic changes in cropland is a vital task to address food safety problems that are essential to zero hunger \cite{muller2015mining}, which can provide timely cropland information to ensure cropland production and food security \cite{garnett2013sustainable}.
% Food security depends on the sustainable management of agricultural land. EO data offer timely insights into crop dynamics, enabling the monitoring of land use, yield, and environmental stressors. Rapid detection of cropland loss—due to urban expansion, infrastructure development, or environmental degradation—is critical to safeguarding food systems \cite{muller2015mining, garnett2013sustainable}.

\begin{figure*}[htbp]
	\centering
	\includegraphics[width=1\linewidth]{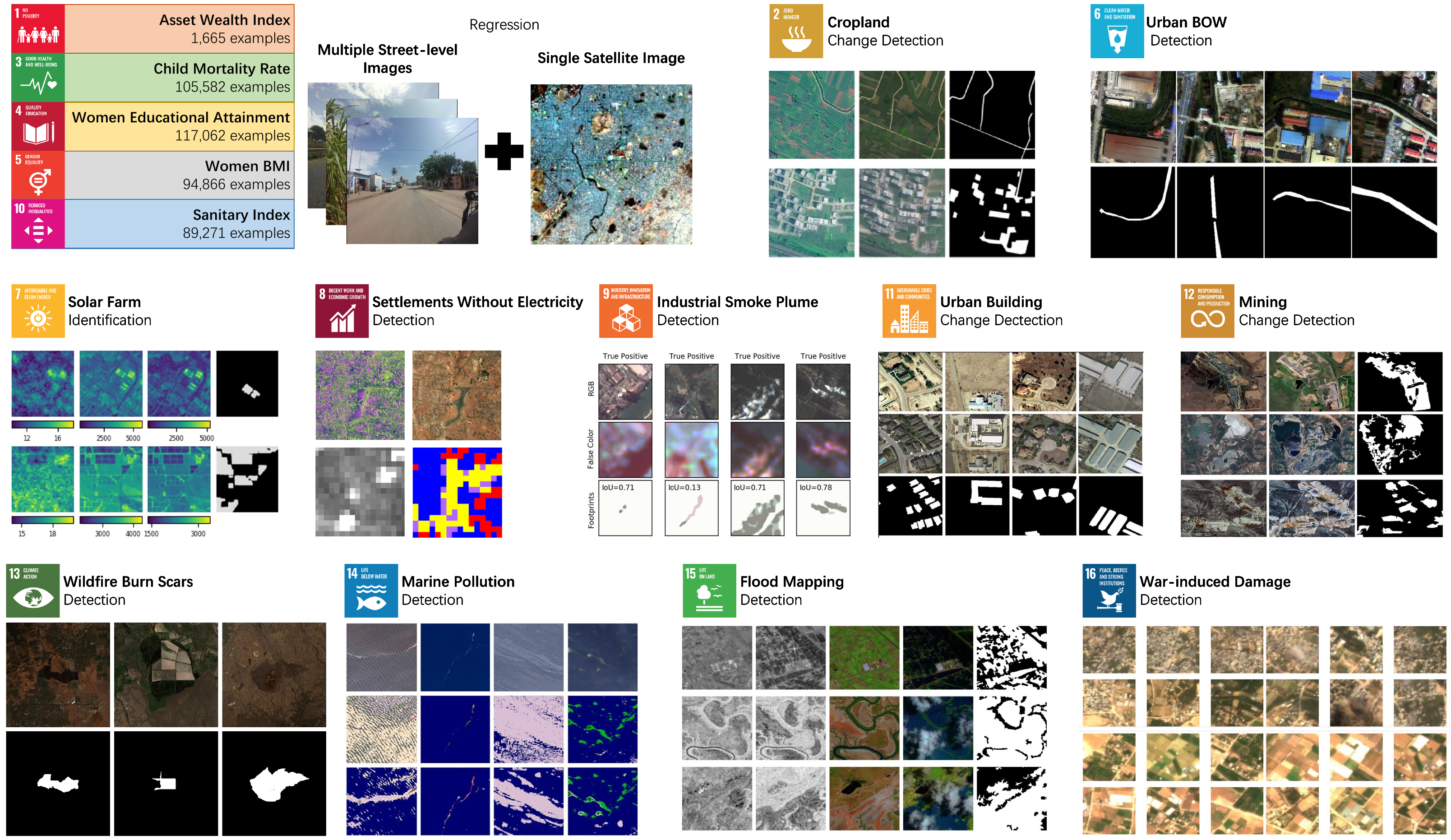}
	\caption{{SDGs tasks and data samples in the SustainFM benchmark.}}
	\label{fig:SGDs_data}
\end{figure*}

SustainFM incorporates the Cropland Change Detection (CLCD) dataset to support food security and sustainable agricultural land management \cite{liu2022cnn}, as shown in Fig.~\ref{fig:SGDs_data}. It comprises 600 cropland change samples derived from the Gaofen-2 satellite in Guangdong Province, China, in 2017 and 2019. Each group of samples consists of two images of $512 \times 512$ pixels and a corresponding binary label indicating the cropland change. Each pair is annotated with binary labels indicating cropland conversion, including changes to buildings, roads, lakes, and bare soil, which are the key factors that threaten agricultural sustainability.

% CLCD comprises 600 paired satellite image samples ($512\times512$ pixels) from Gaofen-2, captured in Guangdong Province, China, between 2017 and 2019. Each pair is annotated with binary labels indicating cropland conversion, including changes to buildings, roads, lakes, and bare soil—key threats to agricultural sustainability.

% \begin{figure}[t]
% 	\centering
% 	\includegraphics[width=1\linewidth]{figures/examples/CLCD.png}
% 	\caption{Samples of the CLCD dataset. It maps the bitemporal changes over time on the cropland area based on Gaofen-2 data.}
% 	\label{fig:CLCD}
% \end{figure}

%The SustainFM involves the CropLand Change Detection (CLCD) \cite{liu2022cnn} dataset that targets monitoring nonagriculturalization incidents that are serious threats to the local agricultural ecosystem and global food security. The CLCD dataset consists of 600 cropland change samples derived from the Gaofen-2 satellite in Guangdong Province, China, in 2017 and 2019, while each group of samples is composed of two images of $512 \times 512$ and a corresponding binary label indicating the cropland change. The main types of change annotated in the CLCD dataset include buildings, roads, lakes and bare soil lands, etc.

\subsection{Good Health and Wellbeing (SDG-3)} %SDG3
Improving health outcomes and ensuring access to healthcare services are vital for enhancing global well-being. 
% Environmental factors such as air quality, water contamination, and heat stress significantly influence public health. 
% EO data offers critical environmental and infrastructural indicators, such as land cover, climate variables, population density, access to clean water, and proximity to healthcare facilities, which are strongly associated with human health outcomes. 
% It can also enable timely, spatially explicit monitoring of public health risks and resource gaps, which significantly benefits governments and organizations in allocating resources more effectively, monitoring progress, and designing targeted strategies to ensure public health. 
Health outcomes are shaped by a complex interplay of environmental and infrastructural factors, including air quality, climate, land use, and access to clean water and healthcare services. EO data offer spatially explicit indicators that can support public health interventions by identifying at-risk populations and monitoring environmental stressors associated with disease and mortality \cite{ford2009using,hu2017spatially,chen2018spatially}. 

The SustainFM involves a task of children's health regression based on the DHS survey dataset of SustainBench \cite{yeh2021sustainbench}. Specifically, the children's health is indicated by the cluster-level child mortality rate that covers children who were age 5 or younger at the time of the DHS survey or who had died (age 5 or younger) no earlier than the year prior to the DHS survey. After identifying the qualifying children, the child mortality rate is further calculated as the number of deaths per $1,000$ children within each cluster. This dataset consists of $105,582$ cluster-level labels for child mortality rates computed from $1,936,904$ children under age 5. 

% SustainFM incorporates a children’s health regression task using DHS survey data from SustainBench \cite{yeh2021sustainbench}. Children’s health is measured via cluster-level child mortality rates, defined as the number of deaths per 1,000 children under age five within a defined spatial unit. The dataset comprises mortality rates for 105,582 clusters, derived from 1,936,904 children surveyed between 1996 and 2019.
%

%The SustainFM involves a task of children's health regression based on the DHS survey dataset of SustainBench. Specifically, the children's health is indicated by the cluster-level child mortality rate that covers children who were age 5 or younger at the time of the DHS survey or who had died (age 5 or younger) no earlier than the year prior to the DHS survey. After identifying the qualifying children, the child mortality rate is further calculated by the number of deaths per 1,000 children by cluster. This dataset consists of 105,582 cluster-level labels for child mortality rates computed from 1,936,904 children under age 5. 

\subsection{Quality Education (SDG-4)} 
Access to quality education is essential for promoting equitable development and reducing social inequalities. Educational access is often influenced by geographical factors, including the proximity of schools, transportation networks, and infrastructure quality. Although EO data cannot directly measure educational outcomes, the triggers for quality education, such as school locations, population density, and education accessibility, are observable on satellite images \cite{runfola2022using,andries2022assessing}. This information helps identify educational disparities, especially in rural and marginalized areas.

% Access to quality education is fundamental to reducing inequalities and fostering sustainable development. Geographical factors such as school proximity, transportation networks, and infrastructure quality significantly influence educational access. While EO data cannot directly measure educational outcomes, they can provide insights into factors affecting education, such as school locations, population density, and regional disparities in educational access \cite{yeh2021sustainbench}. 

% The SustainFM benchmark involves analyzing the average years of women's educational attainment based on the DHS survey dataset of SustainBench. This dataset is obtained based on a survey of women of reproductive age (15-49), and the women’s education metric is created by taking the cluster-level mean of “education in single years” among women between the ages of 15 and 49. The dataset capped the years of education at 18, a common threshold in many surveys, which helps avoid outliers. The dataset contains 122,435 cluster-level labels that were computed from 3,013,286 women across 56 countries.

Therefore, SustainFM includes an analysis of women's educational attainment, based on the DHS survey dataset from SustainBench \cite{yeh2021sustainbench}. The educational attainment metric is derived by calculating the cluster-level mean of education years for women aged 15-49. To mitigate outliers, education years are capped at 18, a common threshold in surveys. Finally, the dataset includes $122,435$ cluster-level labels, computed from data on $3,013,286$ women across 56 countries.

\subsection{Gender Equality (SDG-5)} %SDG5
Achieving gender equality requires eliminating discrimination and ensuring equal access to resources, services, and opportunities. EO-derived indicators such as vegetation cover, urbanization levels, and environmental quality can provide valuable perspectives into the underlying social, economic, and environmental conditions that influence women's health \cite{brown2014using}. By linking these spatial features to BMI outcomes, researchers and policymakers can identify regions where women are particularly vulnerable to malnutrition or obesity due to systemic inequality. This enables the development of geographically targeted, evidence-based interventions aimed at improving women’s health and well-being, thereby addressing one of the critical dimensions of gender inequality. 

The SustainFM involves a task of women's BMI index regression from satellite images based on the DHS dataset of SustainBench \cite{yeh2021sustainbench}. Predicting the BMI from the EO data has demonstrated effectiveness in several previous studies \cite{maharana2018use, dahu2024geospatial}, while the dataset from SustainBench solely focuses on the BMI indexes surveyed from women. In particular, this dataset contains $94,866$ cluster-level BMI labels computed from $1,781,403$ women of childbearing age (15-49), excluding pregnant women. 
%Achieving gender equality requires addressing health disparities rooted in unequal access to resources and services. EO data can reveal spatial patterns linked to these inequalities, such as urban development, vegetation cover, and environmental conditions, which influence women’s health outcomes.
% SustainFM includes a women's body mass index (BMI) regression task, using satellite imagery and survey data from SustainBench \cite{yeh2021sustainbench}. Previous studies have shown the feasibility of predicting BMI from EO data \cite{maharana2018use, dahu2024geospatial}. 
% This dataset includes 94,866 cluster-level BMI labels, derived from 1,781,403 women of reproductive age (15–49), excluding pregnant individuals.

\subsection{Clean Water (SDG-6)} %SDG6
%Access to clean water is critical for public health and sustainable development. Urban water pollution, especially black and odorous water (BOW), poses serious environmental and health risks \cite{cao2020critical}. EO data enables large-scale monitoring of surface water conditions, supporting timely and localized interventions. 
Ensuring universal access to clean water and sanitation is vital for public health and sustainable development. Water scarcity, pollution, and uneven distribution of freshwater resources pose major challenges \cite{cao2020critical}. EO data plays a key role in monitoring surface water extent and quality \cite{modiegi2020comparison,balakrishnan2024remote}. In the residential area, the black and odorous water (BOW) seriously affects the ecological balance of rivers and the health of people living nearby. Therefore, detecting the appearance of such water with EO data helps the authorities with water environment treatment that can improve the water quality in accurate, localized areas. 

SustainFM includes an urban black and odorous water (BOW) detection task using the RSBD dataset \cite{huang2023black}, which comprises 329 multispectral images from the Gaofen-2 satellite in Yantai, China, as shown in Fig.~\ref{fig:SGDs_data}. 
The RSBD dataset is constructed in Yantai, a major port city in the Bohai Sea region in China, which is bordered by the Yellow Sea to the south and the Bohai Sea to the north. Since the northern coastal areas of China have experienced increasing water and energy shortages, as well as offshore environmental pollution, the BOW in the urban area has become a major obstacle to the development of Yantai City, which has limited the sustainable development of the regional economy and society. 
Moreover, pixel-level annotations of BOW enable the training of segmentation models to detect contaminated water, thereby supporting water quality management in urban settings.

\subsection{Affordable and Clean Energy (SDG-7)} %SDG7

Reliable access to clean energy is critical for sustainable development and climate mitigation. With renewables surpassing fossil fuels in several countries (e.g., Germany, for instance, reached a 52\% renewable share in 2024 \cite{german_solar}), tracking the deployment of clean energy infrastructure is increasingly important. EO data enables large-scale identification and monitoring of energy assets, such as solar farms, informing the global transition to low-carbon systems \cite{klingler2024large,bradbury2016distributed,xia2023mapping}.

SustainFM includes a solar farm identification task based on the GloSoFarID dataset \cite{yang2024glosofarid}, which provides $13,703$ multispectral Sentinel-2 image samples with pixel-level annotations. The dataset includes global solar farm samples from 2017 to 2018 and U.S. samples from 2022 to 2023, covering installations at a 10m spatial resolution (shown in Fig.~\ref{fig:SGDs_data}). These EO-derived maps support the analysis of solar infrastructure expansion and guide decision-making in clean energy planning.

\subsection{Decent Work and Economic Growth (SDG-8)} %SDG8
Promoting sustained economic growth and ensuring decent work opportunities are central to reducing poverty and inequalities. Economic development can be indicated by many factors such as urbanization, transportation networks, and infrastructural construction. In modern society, these indicators are significantly influenced by the electricity supply, which is fundamental to everyday life and industrial production \cite{srinivasu2013infrastructure}. However, some regions, especially in less developed countries, still lack an electricity supply, which not only significantly limits economic development but also affects the living quality of local residents \cite{dugoua2018satellite,shah2022electricity}. 

% Sustainable economic growth and equitable work opportunities are key to reducing global poverty. Infrastructure development, such as transportation, housing, and especially access to electricity, is a fundamental enabler of economic activity and quality of life \cite{srinivasu2013infrastructure}. Yet, many regions, particularly in the Global South, still face limited access to electricity, constraining local economies and development. 

The SustainFM involves the task of detecting the settlements without electricity based on the detection of settlements without electricity (DSE) dataset \cite{ma2021outcome}. This dataset was released as part of the 2021 data fusion contest organized by the IEEE Geoscience and Remote Sensing Society (GRSS). The dataset is derived from the Dezda and Salima sectors in Malawi, comprising 98 tiles of multimodal EO data collected from Sentinel-1, Sentinel-2, Landsat 8, and VIIRS nighttime satellites, as shown in Fig.~\ref{fig:SGDs_data}. Each tile is of the size $800 \times 800$ pixels, with each corresponding to a 64 $\mathrm{km}^{2}$ area. The EO data is then mapped into four classes, concerning the existence of human residents and the availability of the electricity supply.

\subsection{Industry, Innovation, and Infrastructure (SDG-9)} %SDG1
Fostering industry, innovation, and infrastructure requires the development of sustainable, resilient systems that are responsive to environmental and societal challenges \cite{costa2024industry}. EO data can support these objectives by providing timely, spatially detailed insights into land use, urban growth, environmental conditions, and infrastructure health. Specifically, EO data enables smarter planning, risk assessment, and innovation in infrastructure design and industrial development, contributing to more sustainable and inclusive industrial growth. For example, detecting smoke plumes through satellite imagery can help monitor industrial emissions, offering valuable information for managing air quality and ensuring the continuity of critical infrastructure \cite{wu2023measurement,sun2023satellite,gomez2024smoke}.

% Sustainable industrial growth hinges on resilient infrastructure and innovation that aligns with environmental and societal goals \cite{costa2024industry}. EO data provides critical insights into land use, urbanization, and environmental change, supporting smarter planning and real-time monitoring. One emerging application is the detection of industrial smoke plumes, a proxy for monitoring emissions and air quality in the context of infrastructure resilience. 

Therefore, the SustainFM dataset involves detecting the industrial smoke plumes from EO data based on the SmokePlume dataset \cite{mommert2020characterization}. The dataset acquires geographic locations of 624 sites from the European Pollutant Release and Transfer Register, which is a pollution reporting entity within the European Union. Subsequently, the dataset retrieved Sentinel-2 satellite imagery taken during 2019 for each site. Each raster image consists of all 12 spectral-band channels from the calibrated Level-2A reflectances and is cropped to a patch with a size of 120×120 pixels, corresponding to an area of $1.2 \times 1.2 \mathrm{km}^2$, as shown in Fig.~\ref{fig:SGDs_data}. The dataset includes 21,350 images with 3,750 positive (a smoke plume is present) and 17,600 negative (no smoke plume is visible) samples. The dataset also generated masks indicating the pixels of smoke plumes for the positive images.

\subsection{Reduced Inequalities (SDG-10)}
Reducing inequality involves addressing income disparities, social exclusion, and unequal access to resources and services. Inequality often manifests spatially, with marginalized communities having poorer infrastructure and fewer services. Unequal access to sanitary services often leaves marginalized groups, such as low-income households, rural populations, and people living in informal settlements, without safe and dignified sanitation and threatens public health. 
EO provides an effective means to delineate these regions and analyze their socio-economic patterns, enabling the estimation of a continuous index that characterizes different degrees of deprivation \cite{persello2020towards}.

SustainFM includes an analysis of the sanitation index (unequal access to sanitary services), based on the DHS survey dataset from SustainBench \cite{yeh2021sustainbench}. The dataset utilizes a cluster-level toilet index to represent the sanitation index, ranked on a $1-5$ scale, where $5$ is the “highest quality”. The dataset comprises a sanitation index for $89,271$ clusters, computed from $2,143,329$ households across 49 countries.

\subsection{Sustainable Cities and Communities (SDG-11)}
As urban populations surge, sustainable city development is crucial for managing infrastructure growth, mitigating environmental degradation, and ensuring equitable living conditions. 
Building construction serves as a key indicator of urban dynamics, reflecting shifts in population, economy, and land use.  Monitoring these changes is essential for managing sustainable urban development. EO-based building change detection enables timely assessment of construction trends, informal settlements, and green space loss, informing urban planning, disaster response, and infrastructure equity \cite{zheng2021building,shen2021s2looking}. 

% SustainFM includes a building change detection task as a key proxy for urban dynamics by leveraging the LEVIRCD dataset \cite{chen2020spatial}. The dataset captures significant urban development through Google Earth high-resolution imagery (0.5 m/pixel) across Texas, USA, spanning 2002–2018. It includes 637 image pairs (1024 × 1024 pixels), annotated with building additions and removals, encompassing diverse structures such as residential villas, high-rise apartments, garages, and warehouses. 

The SustainFM benchmark involves building change detection from remote sensing imagery based on the LEVIRCD dataset \cite{chen2020spatial}. The LEVIRCD dataset focuses on significant land-use changes related to building, especially construction growth, spanning 5 to 14 years from 2002 to 2018 in the Texas region, U.S. It covers various types of buildings, such as villa residences, tall apartments, small garages and large warehouses. The dataset consists of $637$ very high-resolution (VHR, $ 0.5$ m/pixel) Google Earth (GE) image patch pairs, each with a size of $1024 \times 1024$ pixels. Annotations of building additions and removals encompass a diverse range of structures, including residential villas, high-rise apartments, garages, and warehouses.
This supports spatially explicit modeling of urban growth for sustainable city planning.

\subsection{Responsible Consumption and Production (SDG-12)}
Promoting sustainable consumption and production is critical for reducing environmental impacts and conserving resources. Unsustainable practices, such as overextraction, waste mismanagement, and pollution, contribute to environmental degradation, which the EO data can accurately capture. Specifically, land use and land cover changes can provide a timely footprint on human activities, enabling the tracking of their responsibility. Meanwhile, multitemporal EO data offers a powerful means to monitor human activities, allowing for the dynamic tracking of land use changes linked to resource exploitation \cite{li2022change,du2022open,xie2023gan}. 

% utilizing the multitemporal EO data can further dynamically monitor the sustainability of consumption and production activities. 

% Sustainable consumption and production are critical to reducing environmental degradation and conserving natural resources. 

SustainFM addresses environmental degradation and conserves natural resources by monitoring changes caused by mining activities globally, based on the MineNetCD dataset \cite{minenetcd}. As shown in Fig.~\ref{fig:SGDs_data}, MineNetCD includes over $70,000$ bitemporal image patches ($256 \times 256$ pixels) from 100 global mining sites across six continents, collected via Google Earth Engine (GEE). Pixel-level change masks facilitate the assessment of mining expansion, supporting impact assessments and the pursuit of responsible production.

% \begin{figure}[t]
% 	\centering
% 	\includegraphics[width=1\linewidth]{figures/examples/minenetcd.png}
% 	\caption{Samples of the MineNetCD dataset \cite{minenetcd}. It maps the pixels indicating the changes related to the mining development on the Google Earth imagery.}
% 	\label{fig:minenetcd}
% \end{figure}
%The SustainFM dataset involves change detection in mining sites from remote sensing images based on the MineNetCD dataset \cite{minenetcd}. The MineNetCD dataset aims to monitor the expansion of mining sites over a long production period, benefiting the environmental impact assessment in the mining industry to achieve responsible production. The MineNetCD dataset consists of optical bitemporal satellite images collected from 100 global mining sites across six continents, which are derived from the Google Earth Engine (GEE). The MineNetCD dataset crafted a change mask for each bitemporal image pair of mining sites to indicate the changed pixels of the mining area. Finally, the dataset consists of over 70k patches of size $256 \times 256$.

% \begin{figure}[t]
% 	\centering
% 	\includegraphics[width=1\linewidth]{figures/examples/HLSBurns.png}
% 	\caption{Samples of the  HLS Burn Scar Scenes dataset \cite{HLS_Foundation_2023}. It maps burn scars over the contiguous United States from Harmonized Landsat and Sentinel-2 imagery.}
% 	\label{fig:hlsburn}
% \end{figure}

\subsection{Climate Action (SDG-13)}
Addressing climate change requires reducing greenhouse gas emissions, mitigating environmental degradation, and enhancing resilience to extreme weather events. EO data plays a crucial role in tracking climate variables, including land surface temperature, deforestation, and sea-level rise, as well as monitoring hazards such as floods, droughts, and wildfires \cite{tsatsaris2021geoinformation}. In particular, mapping wildfires from satellite imagery provides spatial insight into fire frequency, intensity, and burn extent, informing carbon emission estimates and ecosystem recovery efforts \cite{singh2025active,rashkovetsky2021wildfire}. 
These data-driven insights support disaster preparedness and enable adaptive responses to climate-related risks.

The SustainFM benchmark includes a wildfire impact assessment task using the HLS Burn Scar Scenes dataset \cite{HLS_Foundation_2023}, as shown in Fig.~\ref{fig:SGDs_data}. Based on the U.S. Monitoring Trends in Burn Severity (MTBS) database, this dataset combines burn scar shapefiles with Harmonized Landsat and Sentinel-2 (HLS) imagery acquired one to three months post-fire. 
It is created based on the Monitoring Trends in Burn Severity (MTBS) historical fire database, constructed in the U.S. region, which provides the location of wildfire emergence and the shape file of burn scars.
It consists of 805 scenes, each measuring $512 \times 512$ pixels, with corresponding pixel-wise annotations of burn scars to support precise wildfire detection and post-event analysis.

\subsection{Life below Water (SDG-14)}
Safeguarding marine ecosystems requires urgent action to mitigate pollution and reduce anthropogenic stressors. EO provides a critical means of monitoring ocean health, enabling systematic assessment of sea surface temperature, chlorophyll concentration, turbidity, and harmful algal blooms. Moreover, EO can identify marine pollution ranging from oil slicks to plastic gyres, revealing eutrophication hotspots and floating debris to support early warning systems and targeted interventions \cite{kikaki2024detecting,duarte2023automatic}.

The SustainFM benchmark involves a task of detecting the marine debris and oil spill in the ocean based on the MADOS dataset \cite{kikaki2024mados}, as shown in Fig.~\ref{fig:SGDs_data}. It is composed of multispectral Sentinel-2 data, consisting of 174 scenes captured between 2015 and 2022, with approximately 1.5M annotated pixels, which are globally distributed and collected under various weather conditions. The dataset focuses on a wide range of classes of marine objects related to the pollutants in the form of marine debris and oil spills, which cover $4696$ and $234,568$ pixels. The dataset contains 2803 annotated image patches with a size of $240 \times 240$, each covering an area of $2.4 \times 2.4 \mathrm{km}^2$.

\subsection{Life on Land (SDG-15)}
Protecting terrestrial ecosystems hinges on addressing threats from both natural and human activities.
EO data is indispensable for tracking landscape dynamics, from shifts in vegetation cover and soil moisture to the aftermath of natural disasters \cite{stefanidis2025spatiotemporal}. 
Flood mapping via EO, for instance, enables rapid assessment of inundation impacts on forests, croplands, and habitats, which are vital for risk mitigation, restoration, and land stewardship \cite{iglseder2023potential,potapov2022global,crawford2024biodiversity}.

SustainFM includes a flood detection task to advance disaster response and land management, which is built on the OMBRIA dataset \cite{ombria}, as shown in Fig.~\ref{fig:SGDs_data}. This dataset integrates bitemporal, multimodal satellite data derived from Sentinel-1 SAR and Sentinel-2 multispectral imagery—paired with expert-labeled ground truth from ESA’s Copernicus Emergency Management Service. Covering 23 major flood events worldwide (2017–2021), OMBRIA comprises 3,376 annotated images ($256 \times 256$ px).

\subsection{Peace, Justice, and Strong Institutions (SDG-16)}
Promoting peace, justice, and strong institutions requires protecting civilians, upholding human rights, and ensuring accountability during and after conflict. EO data contributes to these goals by enabling the classification and monitoring of war-induced damage to infrastructure, settlements, and cultural heritage sites \cite{hou2024war}. High-resolution satellite imagery can detect changes such as building destruction, displaced populations, and disrupted land use, providing objective evidence in near real-time \cite{yin2025evaluating}. This information supports humanitarian response, aids in post-conflict reconstruction, and can serve as documentation for legal and human rights investigations.

% \begin{figure}[t]
% 	\centering
% 	\includegraphics[width=1\linewidth]{figures/examples/GAZADeepDAV.png}
% 	\caption{Samples of the GAZADeepDAV dataset \cite{bouabid2024gazadeepdav}. The first and second rows show scenes with damage after the conflict, while the third and fourth rows show scenes without damage collected before the conflict.}
% 	\label{fig:gazadeepdav}
% \end{figure}

% The SustainFM benchmark involves analyzing the war-induced damage based on the GAZADeepDAV dataset \cite{bouabid2024gazadeepdav}. The dataset is constructed based on multispectral PlanetScope satellite imagery with 8 bands and a resolution of 3 m. Specifically, the dataset comprises four satellite images of Gaza before the notable events of October 7, and four images taken after the event in 2023. The dataset consists of 7264 patches for no damage and 6196 patches for damage, each with a size of $256 \times 256$ pixels. 

% Upholding peace and justice amid conflict demands timely, verifiable insights. Earth observation (EO) provides critical, impartial evidence by detecting war-related damage to infrastructure, settlements, and cultural heritage. High-resolution satellite imagery reveals shifts in the urban fabric—collapsed buildings, displaced populations, and disrupted land use—supporting humanitarian response, reconstruction, and legal accountability. 

SustainFM contains a task to detect war-related damage using the GAZADeepDAV dataset \cite{bouabid2024gazadeepdav}, which features eight-band PlanetScope images at $3$m resolution (shown in Fig.~\ref{fig:SGDs_data}. It comprises four pre- and four post-conflict scenes from Gaza in 2023, yielding $13,460$ annotated patches ($256 \times 256$ px), from which $7,264$ were labeled as undamaged and $6,196$ as damaged. GAZADeepDAV provides a distinctive foundation for developing EO tools that enhance transparency and resilience in conflict zones.

\begin{table*}[t]
\caption{Overview of the foundation models for benchmarking.}
\label{table:FMs2}
\centering
% \scriptsize
% \renewcommand{\arraystretch}{1.3}
% \setlength{\tabcolsep}{4pt}
\resizebox{\linewidth}{!}{
\begin{tabular}{l c l l c c}
\toprule
\textbf{Model} & \textbf{Architecture} & \textbf{Pretrained EO Data} & \textbf{Learning Strategies} & \textbf{Parameters (M)} & \textbf{Year} \\
\hline
\rowcolor{gray!10} CROMA & ViT & SSL4EO-S12 \cite{wang2023ssl4eo} & Contrastive & 396.13 & 2023 \\
DOFA & ViT & DOFA \cite{dofa} & MIM & 178.20 & 2024 \\
\rowcolor{gray!10} GFM-Swin & Swin-T & GeoPile \cite{gfmswin2023} & MIM & 128.36 & 2023 \\
Prithvi & ViT & Prithvi-HLS \cite{prithvi} & MIM & 153.28 & 2023 \\
\rowcolor{gray!10} RemoteClip & ViT & SEG-4, DET-10, RET-3 \cite{liu2024remoteclip} & Contrastive & 154.34 & 2024 \\
SatlasNet & Swin-T & SatlasPretrain \cite{satlaspretrain} & Supervised & 128.57 & 2023 \\
\rowcolor{gray!10} ScaleMAE & ViT & FMoW-RGB \cite{christie2018fmow} & MIM & 396.21 & 2023 \\
SpectralGPT & ViT & fMoW-S2 \cite{christie2018fmow}, BigEarthNet \cite{sumbul2019bigearthnet} & MIM & 614.75 & 2024 \\
\rowcolor{gray!10} SSL4EO-S12 & ViT & SSL4EO-S12 \cite{wang2023ssl4eo} & MIM & 61.99 & 2022 \\
SoftCon & ViT & SSL4EO-S12 \cite{wang2023ssl4eo} & Contrastive & 242.19 & 2024 \\
\bottomrule
\end{tabular}}
\end{table*}

%\subsubsection{Partnerships for the Goals (SDG 17)}
%Strengthening global partnerships is essential for achieving the SDGs through cooperation, data sharing, and capacity building. EO data supports cross-border collaborations in monitoring environmental changes, resource management, and disaster response. International data-sharing platforms, such as the Group on Earth Observations (GEO) and Copernicus, enable large-scale monitoring and foster partnerships for sustainable development.
% Specify the data from different sources are provided over the world and covers global area

\definecolor{darkgreen}{RGB}{0,100,0}

% \begin{table*}[t]
% \centering
% \caption{Strengths and weaknesses of FMs (decoder fine-tuned) compared with Traditional Models across performance, efficiency, and sustainability criteria. Color-coded checkmarks indicate relative advantages: \textcolor{darkgreen}{best}, \textcolor{blue}{competitive}, and \textcolor{red}{limitations}.}
% \label{tab:fm_vs_traditional}
% \begin{tabular}{lcc}
% \hline
% \textbf{Assessment Criteria} & \textbf{FMs (Decoder Fine-tuned)} & \textbf{Traditional Models} \\
% \hline
% \rowcolor{gray!10} Model Performance (Accuracy) & \textcolor{darkgreen}{\checkmark\ most tasks} & \textcolor{blue}{\checkmark\ for specific tasks} \\
% Convergence Speed & \textcolor{darkgreen}{\checkmark} & \textcolor{red}{\ding{55}} \\
% \rowcolor{gray!10} Generalization & \textcolor{darkgreen}{\checkmark} & \textcolor{red}{\ding{55}} \\
% Data Efficiency & \textcolor{darkgreen}{\checkmark\ low-shot tuning} & \textcolor{red}{\ding{55}\ requires full training} \\
% \rowcolor{gray!10} Energy Consumption (Training) & \textcolor{darkgreen}{\checkmark} & \textcolor{red}{\ding{55}} \\
% Energy Consumption (Inference) & \textcolor{darkgreen}{\checkmark} & -- \\
% \rowcolor{gray!10} CO\textsubscript{2} Emissions & \textcolor{darkgreen}{\checkmark} & -- \\
% Task Adaptability (SDGs) & \textcolor{darkgreen}{\checkmark} & -- \\
% \hline
% \end{tabular}
% \end{table*}

\section{Geospatial Foundation Models}

To ensure a rigorous and meaningful comparison, we follow PANGAEA \cite{marsocci2024pangaea} to select FMs by emphasizing reproducibility, methodological innovation, and scientific impact. 
Table \ref{table:FMs2} provides a comprehensive overview of the selected models, detailing their architectural scale, the datasets used for pretraining, and the associated EO sensors. These models represent diverse pretraining paradigms commonly employed in geospatial foundation modeling. Specifically, the selected FMs span three major pretraining strategies. (1) Contrastive learning leverages auxiliary data in an unsupervised manner, aligning representations across different modalities to learn robust and transferable features. (2) Masked image modeling (MIM) employs a self-supervised approach, where portions of the input (i.e., patches of an image) are masked and the model is trained to reconstruct the missing content, thereby learning semantically meaningful representations. (3) Supervised learning, by contrast, relies on labeled datasets to train the model directly for specific vision tasks, enabling the model to develop task-relevant perceptual capabilities. Below, we provide a detailed account of the ten FMs selected for evaluation, grouped according to their pretraining strategies: contrastive learning, masked image modeling, and supervised learning. These categories reflect the dominant paradigms currently shaping the development of general-purpose models in EO. 

% An overview of the selected FMs is presented in Table~\ref{table:FMs2}, including their architectural backbones, pretraining datasets, learning strategies, model scales, and publication years.

\subsection{Contrastive Learning}

Contrastive learning leverages auxiliary data in an unsupervised manner, aligning representations across different modalities to learn robust and transferable features \cite{10697182}. This approach has shown particular promise in geospatial applications where multiple data modalities are naturally available.

\textbf{CROMA} \cite{croma} aligns geographically and temporally co-registered SAR-optical image pairs using a contrastive objective, while also incorporating unimodal MIM losses to enhance modality-specific representations. This dual approach enables the model to learn both cross-modal correspondences and modality-specific features, making it particularly effective for applications requiring integration of SAR and optical data.

\textbf{RemoteCLIP} \cite{liu2024remoteclip} bridges the gap between vision and language by aligning satellite imagery with textual descriptions, resulting in visual representations enriched with high-level semantic concepts. By leveraging natural language descriptions, RemoteCLIP enables more intuitive interaction with geospatial data and supports zero-shot classification tasks based on textual queries.

\textbf{SoftCon} \cite{softcon} introduces a multi-label supervision framework to learn cross-scene soft similarities, addressing challenges associated with rigid positive sample definitions and diverse semantic content in EO imagery. This approach recognizes that geospatial scenes often contain multiple land cover types and semantic concepts, requiring more nuanced similarity measures than traditional contrastive methods.

\subsection{Masked Image Modeling}

Masked image modeling employs a self-supervised approach, where portions of the input (i.e., patches of an image) are masked and the model is trained to reconstruct the missing content, thereby learning semantically meaningful representations \cite{10314566,10756791}. This strategy has been widely adopted in geospatial FMs due to its effectiveness in learning spatial patterns without requiring labeled data.

\textbf{DOFA} \cite{dofa} employs dynamic spectral adaptation to pretrain on multimodal EO data, explicitly incorporating physical characteristics of spectral bands based on wavelength information. This physics-informed approach enables DOFA to better understand the relationships between different spectral channels and their physical meaning.

\textbf{GFM-Swin} \cite{gfmswin2023} applies cross-sensor reconstruction to learn generalizable representations across EO modalities. By training on diverse sensor types, GFM-Swin develops robust features that transfer effectively across different imaging platforms and acquisition conditions.

\textbf{Prithvi} \cite{prithvi} is the first FM pretrained on Harmonized Landsat and Sentinel-2 (HLS) data from NASA, integrating temporal dynamics during pretraining to capture seasonality and land cover evolution. This temporal awareness makes Prithvi particularly suitable for applications requiring understanding of seasonal patterns and long-term environmental monitoring.

\textbf{SpectralGPT} \cite{hong2024spectralgpt} adapts MIM for multispectral imagery, incorporating domain-specific mechanisms to accommodate spectral data's unique structure and characteristics. The model leverages the sequential nature of spectral bands to improve reconstruction quality and learn more meaningful spectral representations.

\textbf{ScaleMAE} \cite{scalemae} addresses resolution heterogeneity in satellite data, learning robust representations from multimodal inputs at varying spatial scales. This multi-scale approach enables the model to handle the diverse resolution characteristics common in operational EO applications.

\textbf{SSL4EO-S12} \cite{wang2023ssl4eo} serves as a foundational baseline FM, applying standard MIM techniques to Sentinel-1 and Sentinel-2 data in one of the earliest large-scale geospatial FM pretraining efforts. As one of the pioneering works in this field, SSL4EO-S12 established important benchmarks for subsequent geospatial FM development.

\subsection{Supervised Learning}

Supervised learning, by contrast, relies on labeled datasets to train the model directly for specific vision tasks, enabling the model to develop task-relevant perceptual capabilities. While requiring more annotated data, this approach can lead to highly specialized and effective models for specific application domains \cite{wang2022advancing,10547536}.

\textbf{SatlasNet} \cite{satlaspretrain} constructs a comprehensive multimodal dataset with diverse pretraining tasks, leveraging multi-task supervised learning to build an FM capable of capturing task-generalizable knowledge. By training simultaneously on multiple related tasks, SatlasNet develops rich representations that benefit from the complementary nature of different geospatial analysis objectives.

% Recent literature has witnessed rapid progress in developing foundation models (FMs) tailored to Earth Observation (EO) applications. We adopt the selection criteria proposed by PANGAEA~\cite{marsocci2024pangaea} to ensure a rigorous and meaningful comparison, emphasizing reproducibility, methodological innovation, and scientific impact. Based on these criteria, we include 10 state-of-the-art FMs in our evaluation.

\section{Evaluation}
\subsection{Implementation}
The SustainFM benchmark involves a variety of datasets with diverse spectral features, resolutions, and image sizes, which are designed for different AI applications. This diversity presents challenges in evaluation. First, we preprocessed the dataset into a unified format to facilitate the data loading procedure for fine-tuning and testing the FMs. We resampled all the bands according to the highest resolution for the multispectral data, with different resolutions across the spectral bands. To ensure optimal performance and facilitate the convenient computation of AI algorithms, we crop the large-sized satellite images into small patches of approximately $256 \times 256$. To ensure data transparency and reproducibility, all preprocessed datasets have been uploaded to the Huggingface platform, allowing them to be downloaded and reused within our implementation framework.

FMs adopt the pretraining-fine-tuning paradigm, with their pretrained encoders frozen and decoders fine-tuned on SDG-specific tasks. To ensure broad applicability, we attach task-specific decoders to FM encoders: UperNet for semantic segmentation, Siamese UperNet for change detection, and fully connected layers with global pooling for classification and regression. For the fine-tuning of FMs, cross-entropy loss is used for classification, segmentation, and change detection, while mean squared error (MSE) loss is applied for regression.

Despite the diversity in pretraining strategies, most FMs utilize transformer-based architectures, with a notable consideration of ViT \cite{dosovitskiy2020image}. Therefore, we adopt ViT as a baseline architecture to establish a performance reference. We train ViT from scratch on each subset of the SustainFM benchmark, treating it as a conventional deep learning model. Its performance is then compared against pre-trained and fine-tuned FMs. In addition, we include a traditional deep learning model, ResNet \cite{he2016resnet}, as a representative of convolutional neural networks (CNNs),  widely used for geospatial tasks over the past decade, as the second traditional model.

% We compare various geospatial FMs and traditional deep models using the results obtained on the SustainFM Benchmark. SustainFM encompasses data from six continents and incorporates information from a diverse range of satellite sensors, featuring spatial resolutions ranging from 0.5 to 30 meters. To ensure task diversity, we include a variety of AI tasks such as regression (Reg.), semantic segmentation (Seg.), change detection (CD), and classification (Cls.), all of which are grounded in real-world applications, and include a total of 723,555 image patches with patch sizes ranging from 120×120 to 1024×1024 to reflect reality. 
\subsection{Accuracy Assessment}
Table~\ref{table:FMs3} presents the performance of various FMs (i.e., CROMA \cite{croma}, RemoteCLIP \cite{liu2024remoteclip}, SoftCon \cite{softcon}, DOFA \cite{dofa}, GFM-Swin \cite{gfmswin2023}, Prithvi \cite{prithvi}, SpectralGPT \cite{hong2024spectralgpt}, ScaleMAE \cite{scalemae}, SSL4EO-S12 \cite{wang2023ssl4eo}, and SatlasNet \cite{satlaspretrain}) and traditional deep models (i.e., ViT \cite{dosovitskiy2020image} and Resnet-50 \cite{he2016resnet}) across the SustainFM tasks.
We report performance using the root mean square error (RMSE) for regression and the mean F1 score (mF1) for classification, capturing both the prediction of continuous values and the identification of categorical classes.

% with their pretrained encoders frozen and decoders fine-tuned on SDG-specific tasks, 
It can be seen that FMs generally exhibit superior performance across diverse tasks. This advantage stems from their ability to leverage large-scale pretraining, allowing them to adapt and generalize more effectively across various SDG challenges. Additionally, FMs converge faster in the training stage compared to traditional models, as their pretraining allows them to start with established knowledge, reducing the time and data needed for fine-tuning on SDG tasks. On the other hand, traditional models like ViT and ResNet-50, which are trained from scratch for each SDG task, may require more extensive tuning to achieve optimal performance, especially for broader generalization. 
However, traditional models remain competitive, particularly for task-specific applications where interpretability or fine-tuning on smaller datasets is essential. The decision between FMs and traditional models for SDG-related tasks ultimately depends on factors like model generalizability, the specificity of the task, and available computational resources, but we can highlight the evolving role of pretrained FMs in addressing the complex challenges of sustainable development.
Furthermore, Fig.~\ref{fig:SGDs_data} provides a comprehensive model-wise comparison of the mean F1-score across five different datasets. Among the models, DOFA and GFM-Swin demonstrate strong, consistent performance across most datasets, with their F1-scores frequently reaching the upper range of the chart. The visualization also highlights model-specific strengths and weaknesses, such as the notable performance dip of the Prithvi model on the GloSoFarId dataset (red line), illustrating how a model's effectiveness can vary significantly depending on the data to which it is applied.

\begin{table*}[!htbp]
\scriptsize
\centering
\caption{Performance of Foundation Models and Traditional Methods across SustainFM tasks. Metrics are RMSE (↓) or mF1 (↑) as appropriate. The best performance per column is shown in \textbf{bold}, and the second-best is \underline{underlined}. All results are calculated with $95\%$ confidence interval, using bootstrapped per-sample means (10,000) bootstraps.}
\label{table:FMs3}
\renewcommand{\arraystretch}{1.2}
\setlength{\tabcolsep}{6pt}
\resizebox{\linewidth}{!}{
\begin{tabular}{l|cccccccccccccccc}
\toprule
\textbf{Model} & \textbf{\#1} & \textbf{\#2} & \textbf{\#3} & \textbf{\#4} & \textbf{\#5} & \textbf{\#6} & \textbf{\#7} & \textbf{\#8} & \textbf{\#9} & \textbf{\#10} & \textbf{\#11} & \textbf{\#12} & \textbf{\#13} & \textbf{\#14} & \textbf{\#15} & \textbf{\#16} \\
 & (↓) & (↑) & (↓) & (↓) & (↓) & (↑) & (↑) & (↑) & (↑) & (↓) & (↑) & (↑) & (↑) & (↑) & (↑) & (↑) \\
\hline
CROMA        & 1.46 & 84.02 & 35.01 & 2.75 & 2.10 & 84.94 & \textbf{96.53} & 69.76 & 93.10 & 1.00 & 91.66 & 73.79 & \textbf{91.68} & 70.09 & 87.42 & 96.73 \\
DOFA         & 1.42 & \underline{86.78} & 34.91 & 2.85 & \textbf{2.07} & \underline{88.04} & \underline{96.28} & 72.96 & 93.86 & 1.03 & \underline{94.60} & \textbf{80.70} & \underline{89.03} & \underline{71.08} & \underline{89.26} & \underline{98.15} \\
GFM-Swin     & 1.39 & \textbf{87.69} & \textbf{34.74} & \textbf{2.71} & \underline{2.08} & 87.89 & 94.83 & \textbf{75.82} & 93.58 & 1.04 & 93.78 & 78.06 & 84.55 & \textbf{72.42} & 88.54 & 97.93 \\
Prithvi      & 1.45 & 80.18 & 34.94 & 3.00 & 2.38 & 83.99 & 93.85 & 69.85 & 91.60 & 1.06 & 91.33 & 69.13 & 88.13 & 53.50 & 87.79 & 92.67 \\
RemoteCLIP   & 1.42 & 86.20 & 35.04 & 2.72 & 2.15 & 87.06 & 94.48 & 74.14 & \underline{95.00} & 1.07 & 92.71 & \underline{79.29} & 85.01 & 70.30 & 87.91 & 97.31 \\
SatlasNet    & 1.42 & 82.67 & 34.97 & 2.79 & 2.14 & 82.14 & 94.73 & 70.04 & 93.09 & 1.01 & 92.32 & 73.79 & 87.51 & 63.88 & 87.42 & 96.50 \\
Scale-MAE    & 1.48 & 86.49 & 34.77 & 3.04 & 2.17 & \textbf{88.34} & 95.54 & 69.07 & 94.68 & 1.10 & \textbf{95.00} & 77.16 & 86.03 & 61.39 & \textbf{89.64} & 96.40 \\
SpectralGPT  & \underline{1.39} & 78.97 & \underline{34.75} & \underline{2.71} & 2.16 & 79.29 & 94.70 & \underline{75.27} & 93.91 & 1.04 & 91.41 & 71.73 & 90.07 & 62.54 & 87.58 & 93.12 \\
SSL4EO-S12   & 1.44 & 82.20 & 34.90 & 2.75 & 2.34 & 83.29 & 95.85 & 69.15 & 93.65 & \textbf{0.96} & 91.25 & 70.49 & 88.78 & 64.34 & 87.02 & 95.62 \\
SoftCon      & \textbf{1.39} & 79.97 & 34.79 & 2.74 & 2.13 & 81.34 & 94.94 & 70.02 & 91.16 & \underline{0.98} & 91.56 & 68.03 & 88.03 & 53.37 & 88.12 & 93.98 \\
\midrule
ViT          & 1.50 & 77.72 & 34.81 & 2.89 & 2.20 & 74.29 & 90.58 & 59.91 & 93.17 & 1.10 & 92.73 & 74.52 & 79.17 & 34.56 & 82.61 & 93.17 \\
ResNet-50    & 1.46 & 75.15 & 34.81 & 2.84 & 2.11 & 79.96 & 92.29 & 72.63 & \textbf{95.86} & 1.09 & 92.95 & 75.19 & 81.07 & 56.89 & 82.67 & \textbf{98.17} \\
\bottomrule
\end{tabular}}
\end{table*}

\begin{figure}[!htp]
    \centering
    \includegraphics[width=1\linewidth]{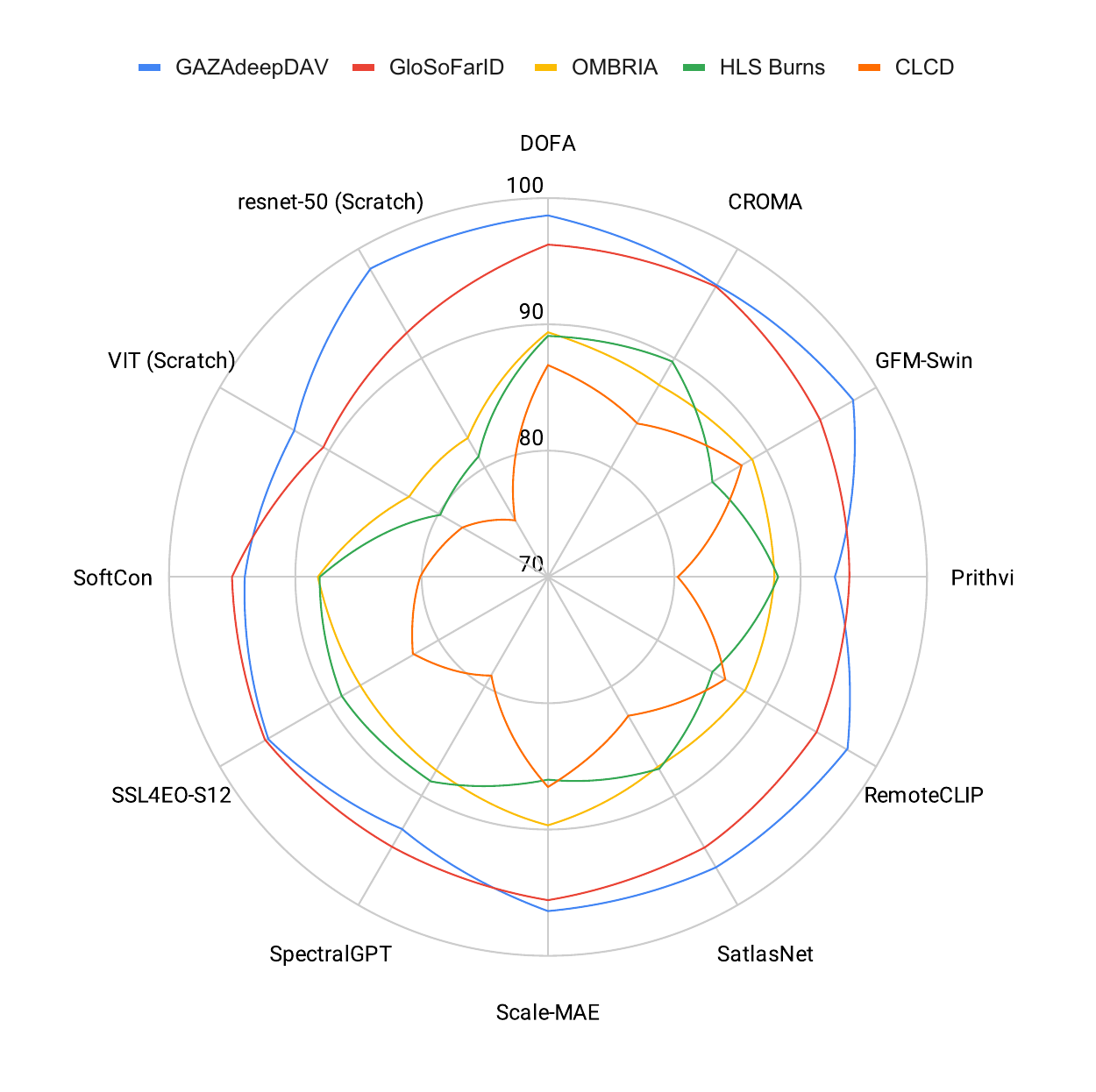}
     % \captionsetup{font=small}
    \caption{Model-wise comparison of mean F1-score across five datasets associated with different SDGs. This visualization highlights the performance differences among evaluated models, and provides insights into their generalization capabilities and robustness.}
    \label{fig:model_compare}
\end{figure}

\begin{table*}[htbp]
% \captionsetup{font=small}
\caption{Comparison of Finetuning the FM decoder only and Finetuning the Full Parameters. ViT and ResNet-50, the representatives of traditional deep learning models, are trained from scratch for SDG 15 (i.e., the OMBRIA dataset for flood mapping).}
\label{table:FMs_fixed_order}
\centering
\scriptsize
\resizebox{\linewidth}{!}{
\begin{tabular}{l|cccc|cccc|c}
\toprule
\multirow{2}{*}{\textbf{Model}} & \multicolumn{4}{c|}{\textbf{Finetuning Decoder Only}} & \multicolumn{4}{c|}{\textbf{Finetuning Full Parameters}} & \multirow{2}{*}{\textbf{CO\textsubscript{2} Difference (\%)}} \\
\cmidrule(lr){2-9}
& \textbf{mF1} $\uparrow$ & \textbf{Training Energy} & \textbf{Inference Energy} & \textbf{CO\textsubscript{2} (kg)}
& \textbf{mF1} $\uparrow$ & \textbf{Training Energy} & \textbf{Inference Energy} & \textbf{CO\textsubscript{2} (kg)} & \\
\midrule

\rowcolor{gray!10} \textbf{CROMA}       & 87.42 & 0.07424 & 0.00118 & 0.04410  & 88.03 & 0.10406 & 0.00122 & 0.06150  & +46.68 \\
\textbf{DOFA}         & 89.26 & 0.03590  & 0.00029 & 0.02121 & 89.46 & 0.05288 & 0.00029 & 0.03111 & +39.46 \\
\rowcolor{gray!10} \textbf{GFM-Swin}    & 88.54 & 0.01766 & 0.00021 & 0.01077 & 89.55 & 0.04023 & 0.00022 & 0.02373 & +120.33 \\
\textbf{Prithvi}      & 87.91 & 0.03473 & 0.00029 & 0.02104 & 87.63 & 0.05246 & 0.00028 & 0.03071 & +61.11 \\
\rowcolor{gray!10} \textbf{RemoteCLIP}  & 87.79 & 0.01497 & 0.00014 & 0.00918  & 87.51 & 0.02500 & 0.00018 & 0.01479 & +61.11 \\
\textbf{SatlasNet}    & 87.34 & 0.03428 & 0.00041 & 0.02028 & 88.97 & 0.09296 & 0.00040 & 0.05436 & +168.05 \\
\rowcolor{gray!10} \textbf{Scale-MAE}   & 89.64 & 0.05802 & 0.00052 & 0.03393 & 89.63 & 0.11246 & 0.00057 & 0.06605 & +94.67 \\
\textbf{SpectralGPT}  & 87.58 & 0.19017 & 0.00142 & 0.11206 & 87.95 & 0.45666 & 0.00160 & 0.26864 & +139.73 \\
\rowcolor{gray!10} \textbf{SSL4EO-S12}  & 87.02 & 0.02355 & 0.00018 & 0.01421 & 87.80 & 0.03371 & 0.00019 & 0.01990 & +40.04 \\
\textbf{SoftCon}      & 88.12 & 0.04170 & 0.00033 & 0.02498 & 88.16 & 0.06883 & 0.00041 & 0.04055 & +62.33 \\
\midrule
\multicolumn{1}{l|}{} & \multicolumn{4}{c|}{} & \multicolumn{4}{c|}{\textbf{Trained from Scratch for traditional deep models}} & \\
\midrule
\rowcolor{gray!10} \textbf{ViT}         & -     & -       & -        & -       & 82.61 & 0.05345 & 0.00031 & 0.03293 & - \\
\textbf{ResNet-50}    & -     & -       & -        & -       & 82.67 & 0.03782 & 0.00011 & 0.02346 & - \\
\bottomrule
\end{tabular}}

\end{table*}

\begin{figure*}[htp]
    \centering

    % Row 1
    \subfloat[]{%
        \includegraphics[width=0.49\textwidth]{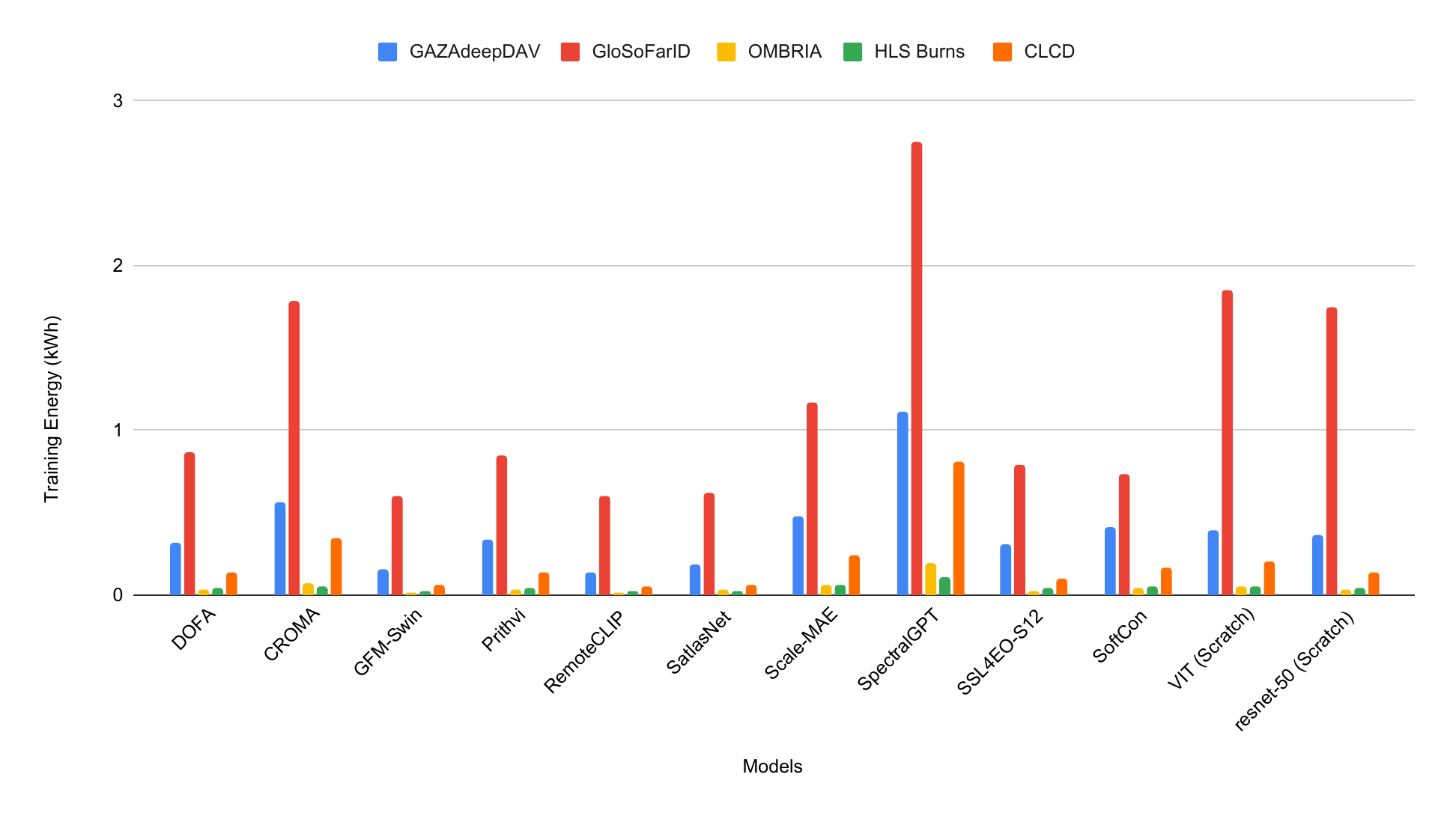}%
    }\hfill
    \subfloat[]{%
        \includegraphics[width=0.49\textwidth]{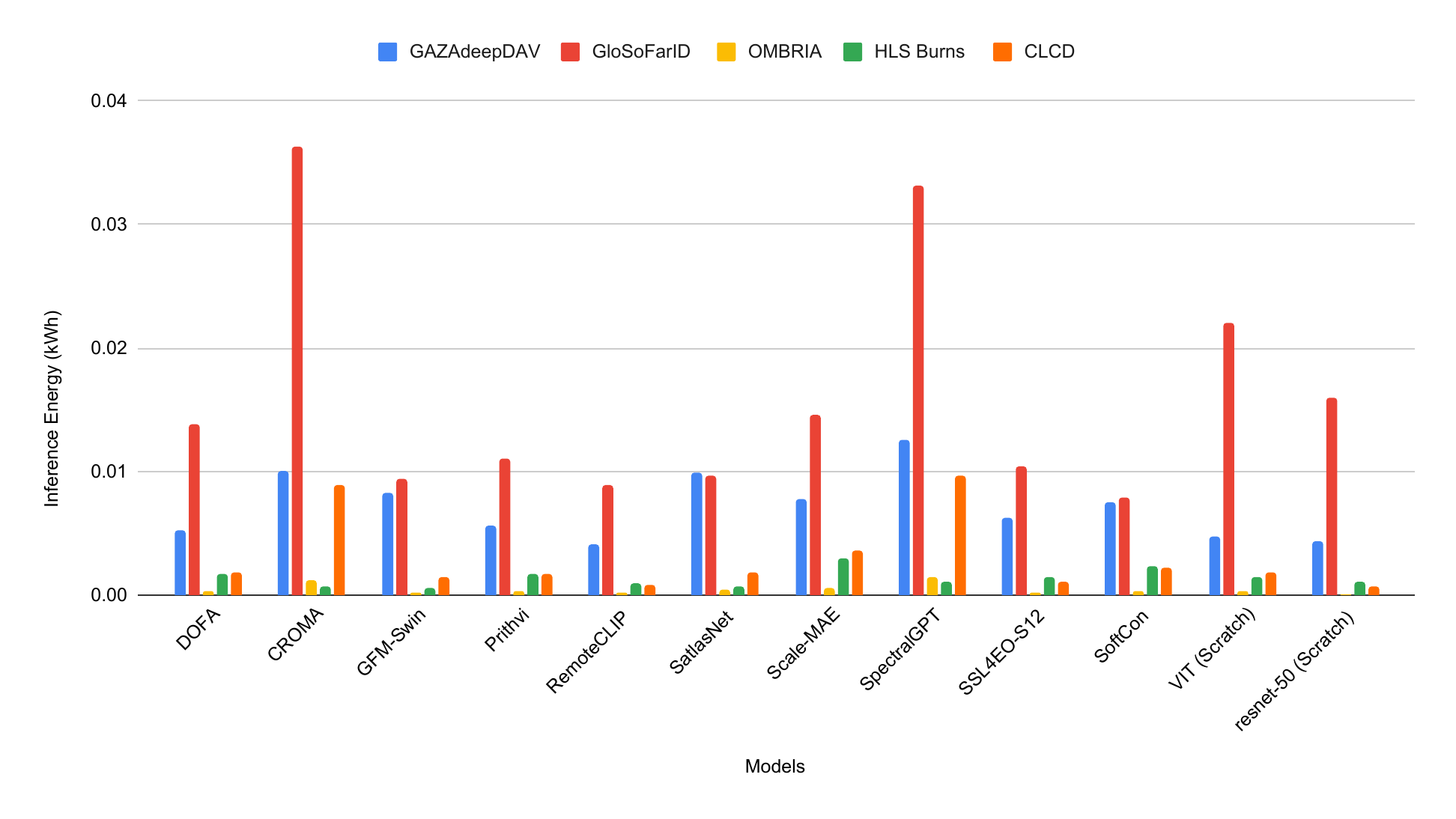}%
    }

    \vspace{-3mm} % 缩小行间距

    % Row 2
    \subfloat[]{%
        \includegraphics[width=0.49\textwidth]{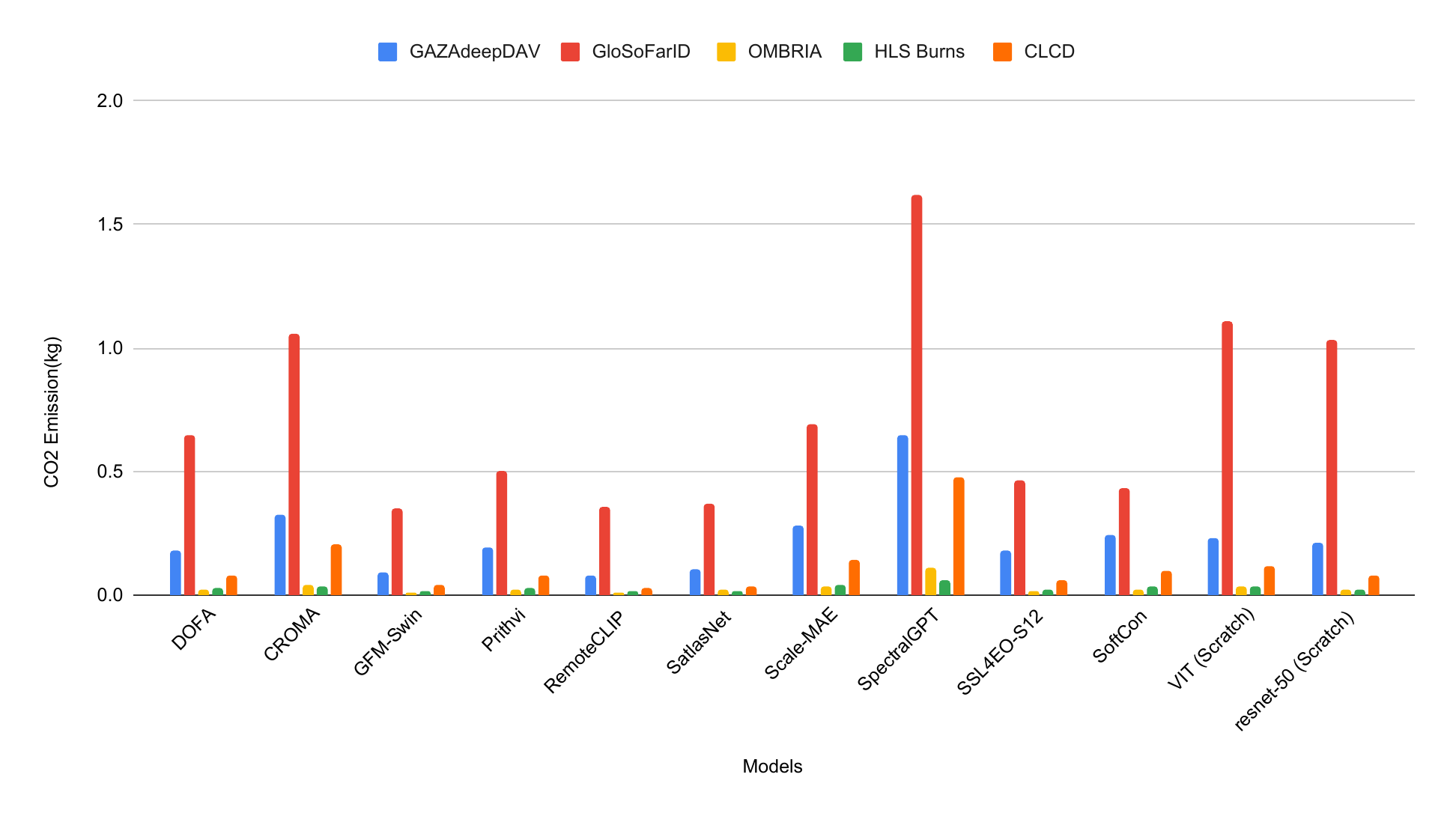}%
    }\hfill
    \subfloat[]{%
        \includegraphics[width=0.49\textwidth]{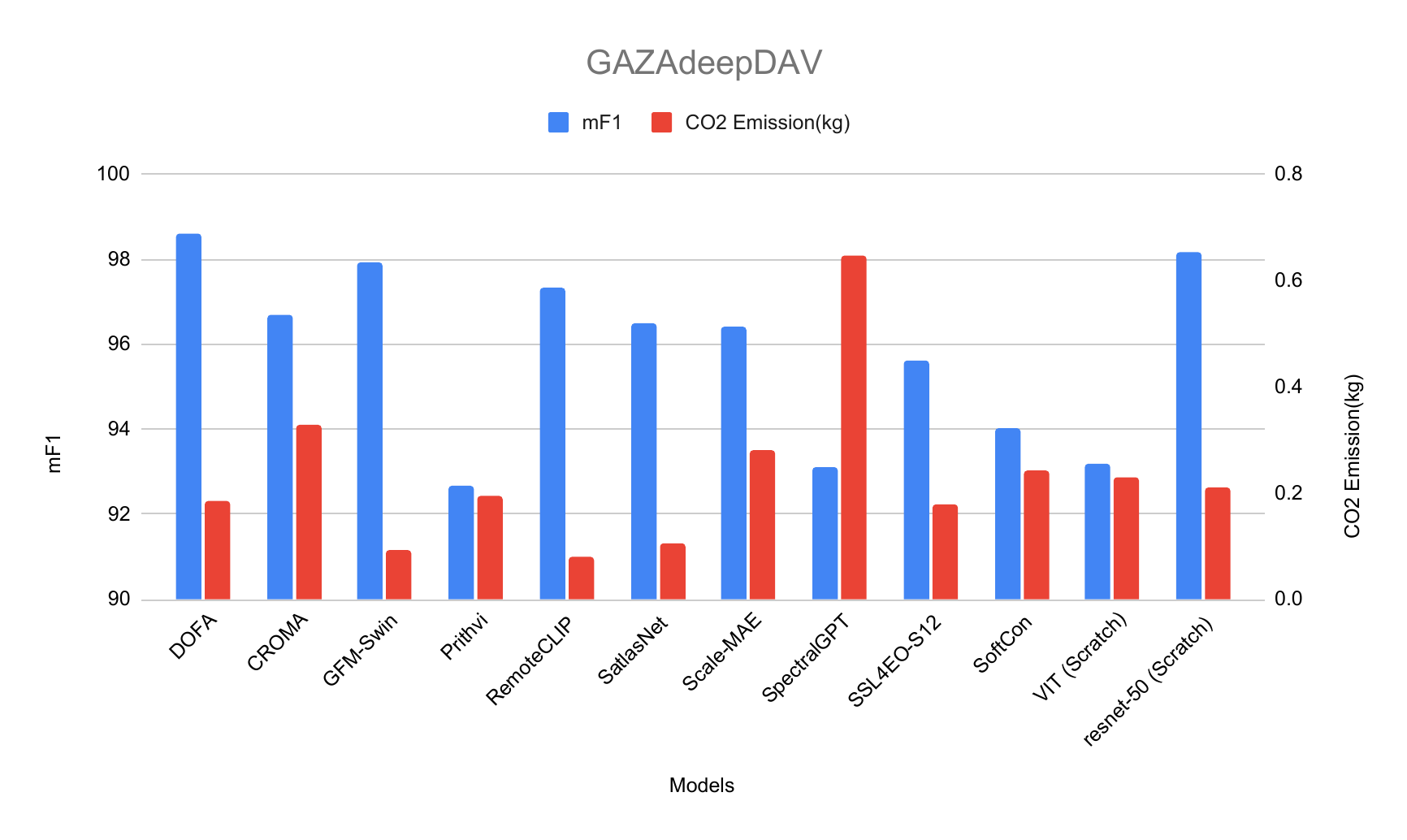}%
    }

    \vspace{-3mm}

    % Row 3
    \subfloat[]{%
        \includegraphics[width=0.49\textwidth]{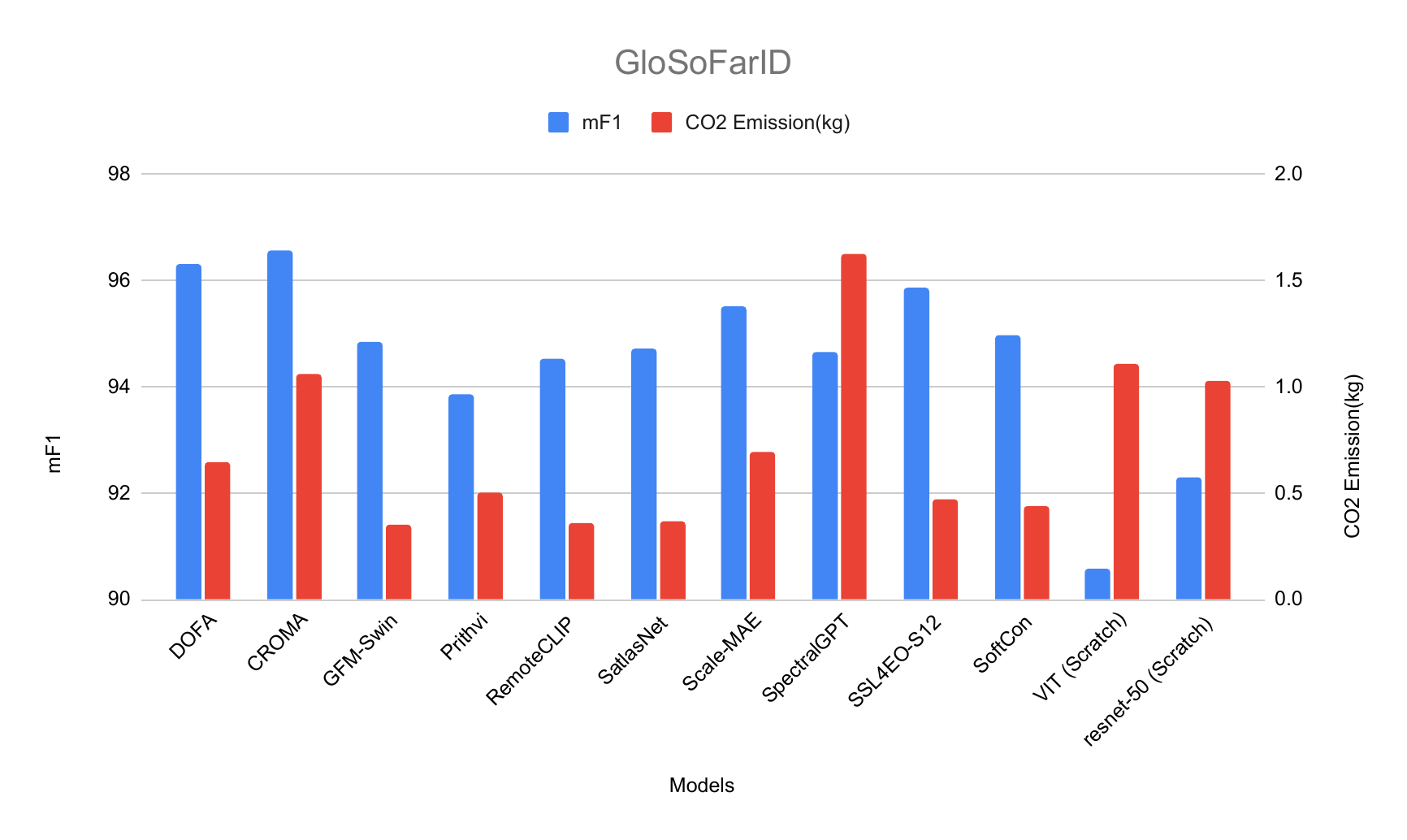}%
    }\hfill
    \subfloat[]{%
        \includegraphics[width=0.49\textwidth]{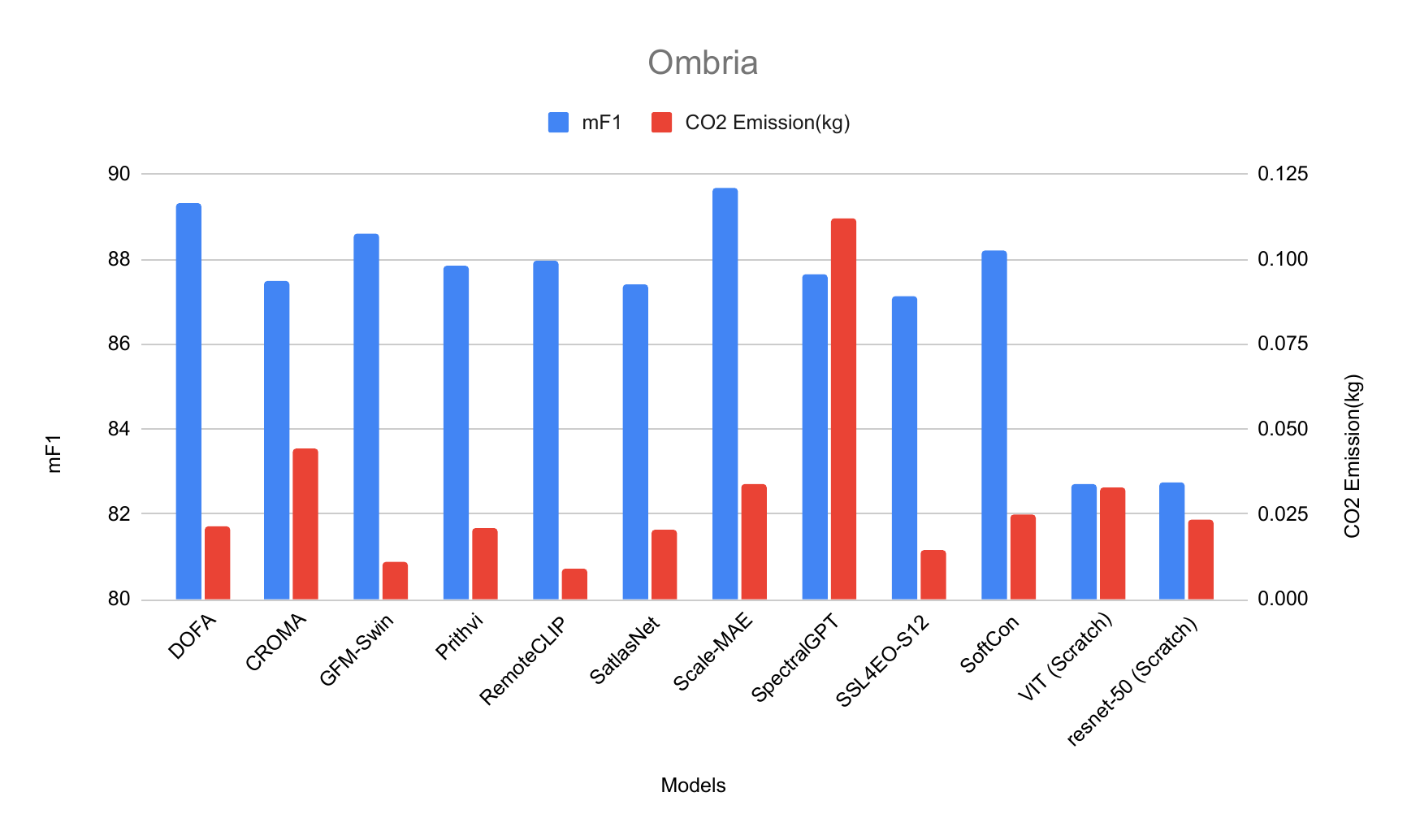}%
    }

    \vspace{-3mm}

    % Row 4
    \subfloat[]{%
        \includegraphics[width=0.49\textwidth]{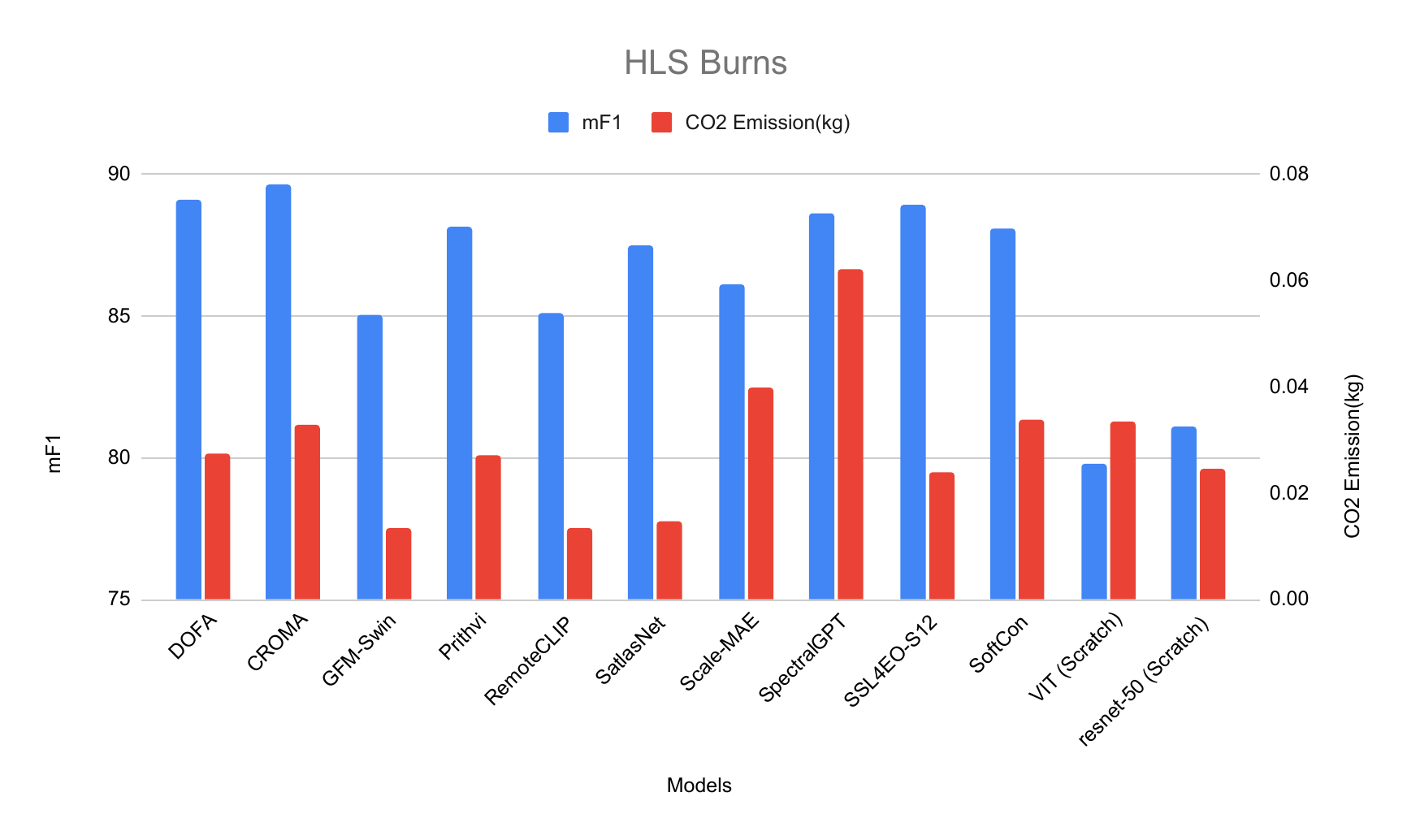}%
    }\hfill
    \subfloat[]{%
        \includegraphics[width=0.49\textwidth]{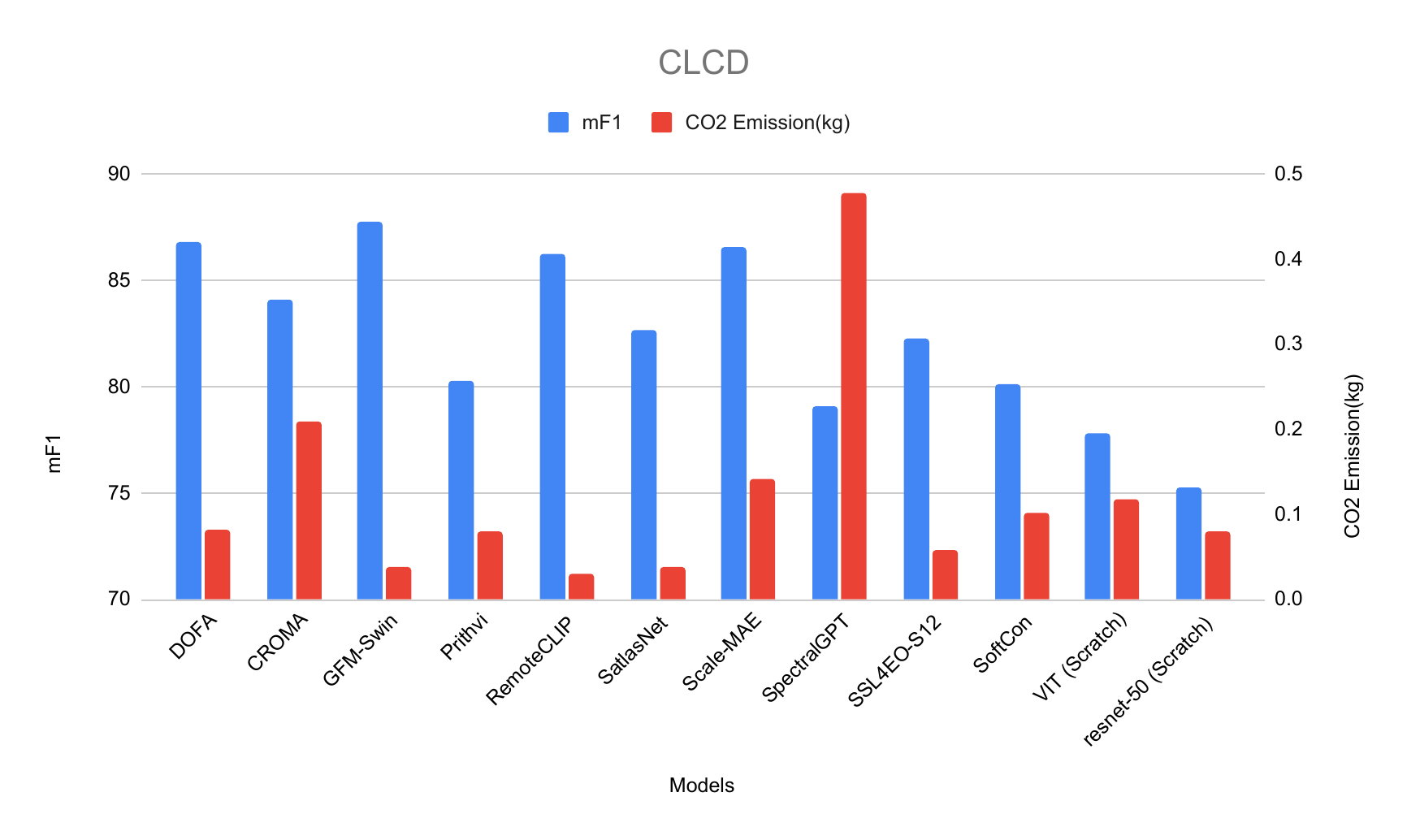}%
    }

    \caption{Comprehensive evaluation of energy efficiency, carbon footprint, and predictive performance across diverse models and datasets. 
    (a) Training energy consumption in kWh, (b) inference energy consumption in kWh, and (c) associated CO\textsubscript{2} emissions in kg across all evaluated models. 
    (d–h) Joint comparison of mF1 and CO\textsubscript{2} emissions for five representative datasets (GAZAdeepDAV, GloSoForID, Ombria, HLS Burns, and CLCD), illustrating the trade-off between model accuracy and environmental sustainability.}
    \label{fig:energy_mf1_comparison}
\end{figure*}

 % Figs.~\ref{fig:model_compare},~\ref{fig:energy_mf1_comparison} and~\ref{fig:reduced_samples_compare}

\subsection{Sustainability-Oriented Assessment}
To assess different models' environmental impact and sustainability, we employ the CodeCarbon library \cite{codecarbon} to monitor energy consumption during training and inference, and show the results in Table \ref{table:FMs_fixed_order}. Specifically, we compare the energy footprint of decoder-only fine-tuning with full model fine-tuning, alongside the energy demands of baseline and traditional architectures (ViT and ResNet-50) trained from scratch. %Table \ref{table:FMs_fixed_order} compares the performance and resource consumption of FMs fine-tuned on the decoder only or with full parameters, together with traditional deep learning models (ViT and ResNet-50) trained from scratch. 
All models are evaluated on the OMBRIA dataset (Life on Land, SDG-15). In general, fine-tuning only the decoder of an FM can slightly boost performance while using less energy and producing lower CO\textsubscript{2} emissions than fine-tuning the entire model. This is because it requires fewer gradient computations and backpropagation steps. This relative difference varies substantially across all models, ranging from +40\% to as much as +168\%, as can be seen in the last column of Table \ref{table:FMs_fixed_order}, highlighting the considerable environmental benefits of updating only the decoder. For context, full fine-tuning of SpectralGPT produces around 270g of CO\textsubscript{2} emissions. This is roughly equivalent to driving a car for 0.7 miles, or less than 1\% of the average American’s daily per-person emissions (39 kg/day) \cite{epa_vehicle_emissions}. Complete training for traditional deep learning models usually results in higher energy consumption and CO\textsubscript{2} emissions compared to partially fine-tuned FMs, since conventional models require training from scratch on large datasets over many iterations. In contrast, FMs benefit from pretraining on diverse data; They can be adapted to downstream tasks using a small amount of labeled data and fewer updates, which makes them more energy-efficient and scalable in the long run.

% Table \ref{table:FMs_fixed_order} compares the performance and resource consumption of FMs fine-tuned on the decoder only or with full parameters, together with traditional deep learning models (ViT and ResNet-50) trained from scratch. All models are evaluated on the OMBRIA dataset (Life on Land, SDG-15). In general, fine-tuning the decoder part of FMs alone leads to slightly better performance with lower training and inference energy consumption and CO\textsubscript{2} emissions, in contrast to fine-tuning the entire network. This is because decoder-only fine-tuning updates significantly fewer parameters, which reduces the number of gradient computations and backpropagation steps required. Full training from scratch for traditional deep learning models usually results in higher energy consumption and CO\textsubscript{2} emissions compared to partially fine-tuned FMs, since traditional models require training from scratch on large datasets over many iterations. In contrast, FMs benefit from pretraining on diverse data and can be adapted to downstream tasks using a small amount of labeled data and fewer updates, which make them more energy-efficient and scalable in the long run.

For a fair comparison across different methods and datasets, Fig.~\ref{fig:energy_mf1_comparison} presents a comprehensive evaluation of models along three key dimensions: predictive performance, energy consumption, and carbon footprint. The analysis reveals a significant trade-off between model accuracy and environmental impact, which is heavily influenced by the chosen dataset. Notably, the GloSoFarId dataset consistently requires the most energy for both training and inference, leading to CO$_2$ emissions that are an order of magnitude higher than all other datasets. While some models, such as SpectralGPT, achieve competitive performance at a substantial environmental cost, others, like DOFA and GFM-Swin, demonstrate a more favorable balance of high mF1 scores and lower carbon footprints. 
High emissions are also observed for the CROMA and ViT (Scratch) models on the same dataset. In contrast, the HLS Burns and CLCD datasets result in significantly lower emissions across all models. 
This work highlights the critical need to expand model evaluation frameworks beyond traditional performance metrics to include the full energy and environmental costs.

\begin{figure}[htp]
    \centering

    % 子图 (a)
    \subfloat[]{%
        \includegraphics[width=\linewidth]{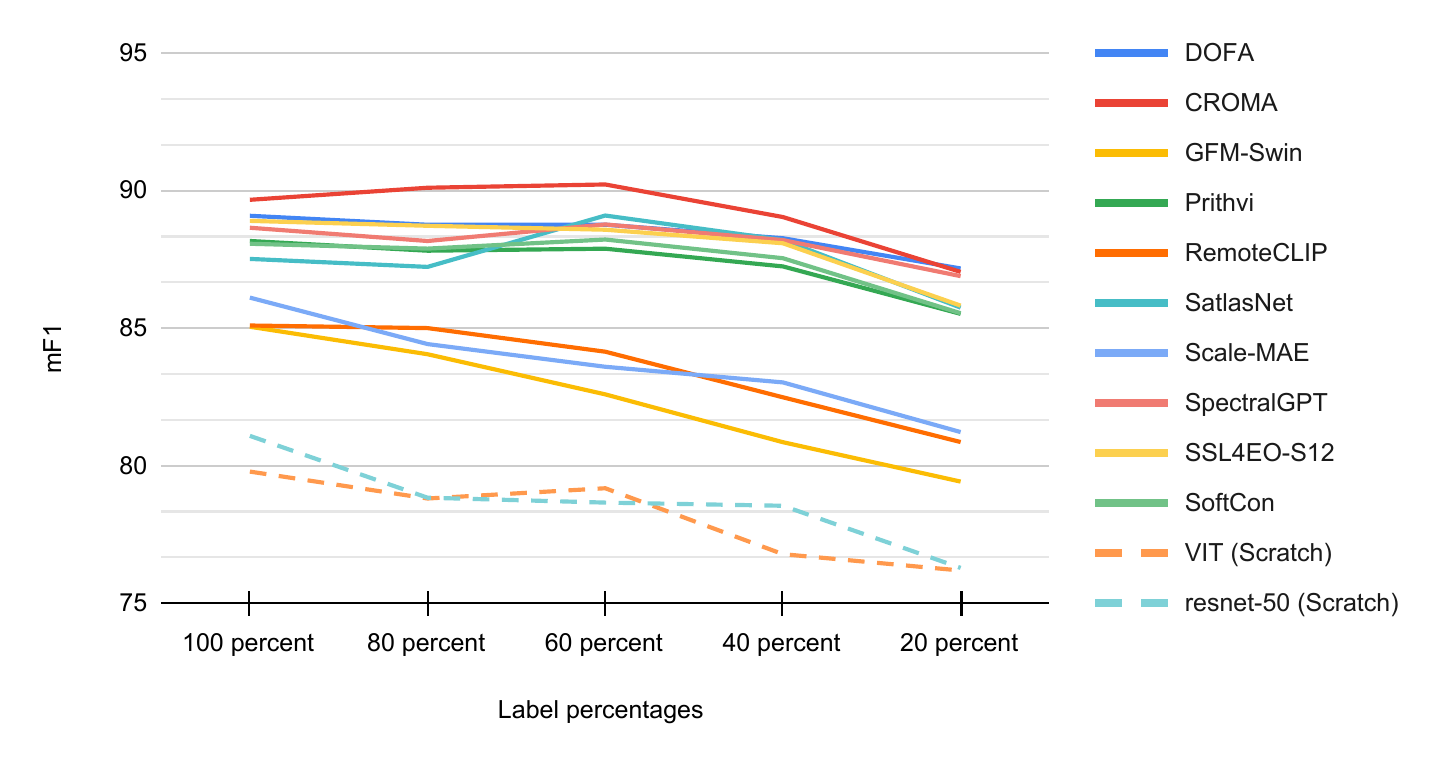}%
    }\\[1em] % 换行并增加间距

    % 子图 (b)
    \subfloat[]{%
        \includegraphics[width=\linewidth]{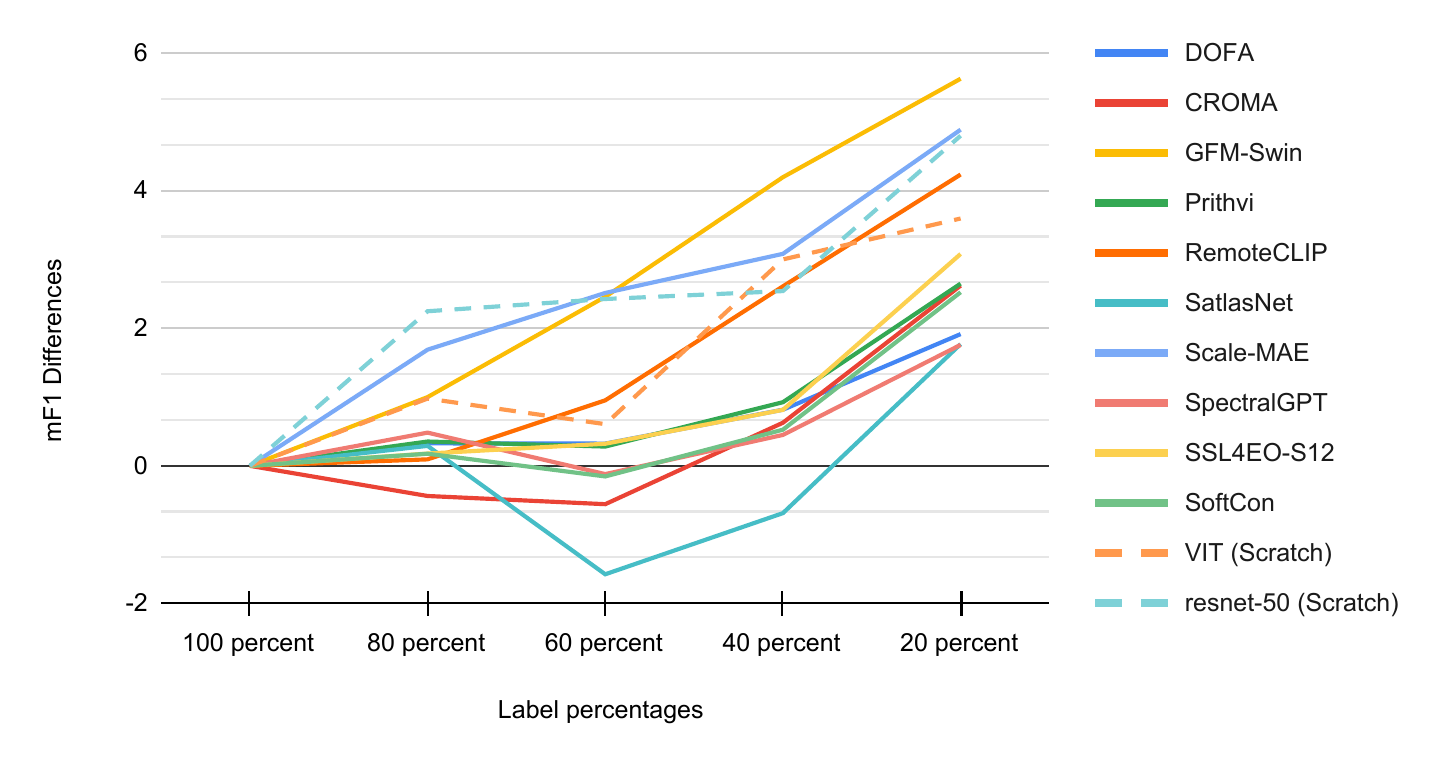}%
    }

    \caption{Impact of training data reduction on model performance for HLS Burns.  
    (a) mF1 scores achieved by each model when trained on progressively reduced subsets of the training data, compared to full training (100\%).  
    (b) Absolute differences in mF1 scores between full training and reduced-sample settings, highlighting model robustness to limited labeled data.}
    \label{fig:reduced_samples_compare}
\end{figure}

Fig.~\ref{fig:reduced_samples_compare} illustrates the impact of reducing training data on the performance of various models when applied to the HLS Burns dataset.
% Fig.~\ref{}(a) plots the mF1 scores of each model 
As the percentage of labeled training data is progressively reduced from 100$\%$ down to 20$\%$, a general downward trend exists for all models, indicating that a decrease in training data leads to a decline in performance. Models like GFM-Swin and DOFA maintain relatively high scores, while others like ViT (Scratch) and resnet-50 (Scratch) experience a more pronounced performance drop.
Fig.~\ref{fig:reduced_samples_compare}(b) provides a more focused view of model robustness by plotting the absolute difference in mF1 score relative to the full (100$\%$) training set. 
It reveals that GFM-Swin and DOFA are the most robust models, as their mF1 scores decrease only slightly. Conversely, models like SpectralGPT and Scale-MAE show the largest performance degradation, indicating that they are more sensitive to the amount of labeled training data.

\begin{table*}
\centering
% \captionsetup{font=small}
\caption{Comparison between Foundation Models (decoder fine-tuned) and Traditional Deep Models (trained from scratch), across performance, efficiency, and sustainability metrics. Color-coded markers indicate relative performance: \textcolor{darkgreen}{\checkmark best}, \textcolor{blue}{\checkmark competitive}, and \textcolor{red}{\ding{55} limitation}.}
\label{tab:fm_vs_traditional}
\renewcommand{\arraystretch}{1.5}
\begin{tabular}{@{}lcc@{}}
\toprule
\textbf{Assessment Criteria} & \textbf{FMs (Decoder Fine-tuned)} & \textbf{Traditional Models} \\
\midrule
\rowcolor{gray!10} Model Performance (mF1)      & \textcolor{darkgreen}{\checkmark high across models}         & \textcolor{red}{\ding{55} lower performance} \\
Training Efficiency (kWh)        & \textcolor{darkgreen}{\checkmark lower training cost}       & \textcolor{blue}{\checkmark moderate} \\
\rowcolor{gray!10} Inference Efficiency (kWh/sample) & \textcolor{darkgreen}{\checkmark low in most cases}       & \textcolor{blue}{\checkmark competitive (ResNet-50)} \\
CO\textsubscript{2} Emissions (kg)            & \textcolor{darkgreen}{\checkmark generally low}             & \textcolor{blue}{\checkmark moderate} \\
\rowcolor{gray!10} Data Efficiency             & \textcolor{darkgreen}{\checkmark few-shot fine-tuning}     & \textcolor{red}{\ding{55} requires full retraining} \\
Generalization across tasks       & \textcolor{darkgreen}{\checkmark broad task adaptability}   & \textcolor{red}{\ding{55} narrower scope} \\
\rowcolor{gray!10} Convergence Speed           & \textcolor{darkgreen}{\checkmark faster, low compute}       & \textcolor{red}{\ding{55} slower convergence} \\
\bottomrule
\end{tabular}
\end{table*}

\section{Discussions}\label{sec5}
Table~\ref{tab:fm_vs_traditional} summarizes the strengths and weaknesses of FMs (decoder fine-tuned) compared to traditional models in terms of performance, efficiency, and sustainability criteria. Here, we discuss our key findings, as well as principles and considerations for geospatial
FMs. Overall, foundation models demonstrate clear advantages in terms of model performance, data efficiency, convergence speed, and generalization capability. They achieve higher mF1 scores with substantially lower training costs and CO\textsubscript{2} emissions, reflecting both computational and environmental benefits. Moreover, their few-shot fine-tuning capability enables rapid adaptation to diverse downstream tasks, contrasting with the full retraining requirements of traditional models. While traditional models such as ResNet-50 remain competitive in inference efficiency, their overall scope and scalability are limited. These results highlight the superior efficiency and adaptability of foundation models for sustainable and general-purpose Earth observation tasks.

\subsection{Key Findings}
\begin{itemize}
\item \textbf{FMs deliver strong performance, but are not always superior to traditional approaches.} Our results indicate that it is impossible to make a universal claim about the superiority of FMs over traditional deep learning models (e.g., ViT) in terms of accuracy. Performance varies depending on the dataset, task, and fine-tuning strategy. Nonetheless, in many of our experiments, FMs outperformed their traditional counterparts.

\item  \textbf{Transferability offers a step toward energy-efficient AI across domains.} A commonly held belief is that FMs are inherently resource-intensive. However, our results suggest a more balanced picture. Once pretrained, FMs can be fine-tuned with minimal labeled data, which offers an energy-efficient alternative to training models from scratch. This transferability makes them especially attractive for zero-shot or few-shot learning scenarios, where resource constraints or data scarcity are significant limitations.

\item  \textbf{FMs enable scalable solutions for global sustainability.} One of the most promising aspects of FMs is their potential to serve as general-purpose tools across geospatial applications. Their ability to generalize across regions, tasks, and modalities makes them particularly suited to addressing complex, large-scale problems related to the SDGs. This cross-domain flexibility allows broader, interdisciplinary adoption among scientists and practitioners.

\item \textbf{High generalization and fast convergence make FMs a scalable alternative to traditional models.} Traditional deep learning models typically require task-specific training from scratch to achieve competitive performance, which results in considerable computational and energy costs for each new use case. In contrast, FMs exhibit strong generalization capabilities across various tasks, often requiring only minimal fine-tuning. This enables faster convergence and lower overall resource consumption in the long term.
\end{itemize}

\subsection{Best Practices}%: What Truly Matters for Geospatial Foundation Models
\begin{itemize}

\item \textbf{Energy efficiency and transparency in FM training:} Training FMs from scratch requires an enormous amount of time and computational resources, which make it largely impractical for benchmarking studies like ours. Given these constraints, we focused on fine-tuning existing pre-trained models. Accordingly, the energy use reported in our work reflects only this fine-tuning phase. We strongly encourage researchers to clearly report CO$_2$ emissions or other environmental impact measures whenever they introduce new FMs to provide the whole picture. This transparency is essential to support responsible AI development and help the community move toward more sustainable practices in training and using large models.

\item  \textbf{Aligning models with the nature of EO:} The development of geospatial FMs demands a methodological shift that reflects the unique nature of EO data. As argued in \cite{10.5555/3692070.3693807}, EO data is not just another visual modality; They include unique properties, such as spectral and temporal richness, geolocation, physical constraints, and sensor-specific noise patterns, which are not adequately addressed by conventional vision-based pretraining objectives such as masked image modeling or contrastive learning or the models themselves \cite{ren2024geospatial}. As a result, downstream geospatial tasks such as change detection, time-series forecasting, and biophysical retrieval may suffer from a misalignment between model pretraining and real-world task demands. Moreover, relying on standard computer vision architectures, such as UperNet or Siamese UperNet, can hide the real impact of task-specific reasoning in geospatial applications. To move the field forward, we advocate for models that capture the specific nature of geospatial data. This includes using physics-informed pretraining that brings in domain knowledge, learning methods that combine different types of data (like spatial, spectral, and temporal), and flexible decoder designs with suitable loss functions that match the needs of specific geospatial tasks.

\item \textbf{Scaling laws for geospatial FMs:} Furthermore, the importance of scale must not be overlooked. Though originally developed for language models, scaling laws emphasize the importance of maintaining an appropriate data-to-parameter ratio, typically ranging from 5:1 to 20:1, to achieve generalizable performance \cite{10.5555/3692070.3693840}. As highlighted in \cite{ren2024geospatial}, many geospatial FMs are undertrained relative to their size, limiting their generalization capacity despite architectural scale. Smaller models risk underfitting complex Earth patterns, while larger models require correspondingly large and diverse training datasets to fulfill their potential.

\item \textbf{Responsible practices with building fair and sustainable models:} The environmental and ethical responsibilities tied to geospatial FM development are substantial. The carbon footprint of model training, particularly for large-scale architectures, remains a concern, especially when models are trained without transparent reporting. Moreover, EO datasets often contain geographic, sensor-based, and socioeconomic imbalances, which can propagate biases into model predictions \cite{ghamisi2025responsible}. These biases risk producing results that are not representative of the true situation. Mitigation strategies include careful data curation, geographic rebalancing, and inclusive benchmarking practices. 

\item  \textbf{Beyond benchmarking toward real-world impact:} Last but not least, we want to emphasize that while technical benchmarking is a necessary step, it is not a sufficient one. Public perceptions significantly influence policy decisions by shaping political priorities, public support, and the feasibility of implementation. Performance metrics alone do not guarantee real-world relevance or societal impact. While benchmarks offer a standardized way to measure methodological progress, they must be grounded in broader, mission-driven goals. As aligned with studies such as \cite{kshirsagar2021goodaigood}, the impactful deployment of geospatial FMs, especially in support of SDG-17, relies on collaboration with stakeholders, domain experts, and decision-makers who can translate model outputs into actionable insights for "Becoming Good at AI for Good". In practice, benchmarking and modeling represent just two components of a complete solution. Defining relevant problems, selecting appropriate metrics, involving domain experts, and clearly articulating the model’s purpose are equally important. FMs, unlike narrow models that suffer from “catastrophic forgetting”, offer unique advantages for long-term monitoring and adaptation to temporal shifts in the Earth system.  Well-pretrained FMs enable lifelong learning and can generalize to new tasks with minimal further tuning.
    
\end{itemize}

The regulation of FMs varies globally. The EU's AI Act (2024) introduces the first binding framework focused on risk management and transparency, classifying large models ($>10^{25}$ FLOPs) as systemically risky. The U.S. Executive Order 14110 (2023) adopts a flexible, security-oriented approach, while China’s 2023 Measures emphasize content labeling and safety reviews. The UK favors principles-based innovation, and Canada’s AIDA provides a balanced model with targeted obligations for high-impact systems.

Together, these differing regulatory approaches illustrate both the complexity and the promise of global AI governance. The divergence is not only a legal matter, but it also influences where innovation occurs and who benefits from it. Yet, within this diversity lies a chance to build a more balanced and trustworthy global AI ecosystem. Through open dialogue, shared standards, and mutual learning, countries can turn regulatory variation into a driver of collective progress rather than fragmentation.

If shaped by cooperation, transparency, and ethical responsibility, emerging governance frameworks can strengthen public trust, improve data stewardship, and support international collaboration. To realize the potential of foundation models for the SDGs (from equitable access to education and healthcare to climate resilience and social inclusion), governance must evolve toward systems that are interoperable, transparent, and human-centered. Aligning oversight with shared principles of openness, accountability, and social benefit can ensure that the next generation of foundation models becomes not only a technological advance, but also a genuine force for sustainable and inclusive global progress.

\section{Conclusion}

In this study, we establish a comprehensive benchmark that integrates mainstream Foundation Models with multi-source datasets aligned with SDG objectives, enabling systematic evaluation of their true societal value and practical utility in supporting global sustainable development. 
Through systematic evaluation across key criteria, including transferability, data efficiency, energy costs, and operational impact, we provide a grounded, rigorous assessment that challenges assumptions and clarifies the actual added value, if any, of FMs in this domain. By moving beyond traditional accuracy metrics to include transferability, energy efficiency, and operational impact, this work addresses the urgent need for AI solutions that demonstrably contribute to global sustainability. Ultimately, our goal is to contribute to global sustainability objectives and promote social good, a key aspect that has been undermined in many current geospatial FMs. 

Geospatial FMs can make EO more accessible, actionable, and relevant by combining innovative methodologies with impactful applications and deep domain expertise. Future FMs will help to address global challenges and advance the SDGs. FMs can deliver actionable insights for sustainable development, humanitarian response, and environmental conservation if designed responsibly, i.e., with attention to energy efficiency, open-access datasets, and a strong emphasis on transparency and inclusivity—geospatial. By democratizing access to AI resources (e.g., input data, curated datasets, model architectures, and pretrained weights), geospatial FMs can reduce disparities in dataset availability and empower stakeholders as a key driver of innovation to foster global progress toward sustainability.

\bibliographystyle{IEEEtran}
\bibliography{Ref} 

\end{document}